\documentclass[letterpaper, 10 pt, conference]{ieeeconf}  

\IEEEoverridecommandlockouts                              

\usepackage{graphics} 
\usepackage{epsfig} 
\usepackage{mathptmx} 
\usepackage{times} 
\usepackage{amsmath} 
\usepackage{amssymb}  

\usepackage{url}
\usepackage{xspace}
\newcommand{\etal}{\emph{et al.}\xspace}

\DeclareMathOperator*{\argmin}{arg\,min}

\usepackage{cite}
\usepackage{multirow}

\usepackage{arydshln}

\usepackage{textcomp}

\usepackage{lipsum}

\usepackage[maxfloats=256]{morefloats}
\title{\LARGE \bf
4D Attention: Comprehensive Framework\\ for Spatio-Temporal Gaze Mapping
}

\author{Shuji Oishi$^{1}$, Kenji Koide$^{1}$, Masashi Yokozuka$^{1}$, Atsuhiko Banno$^{1}$
\thanks{*This work was supported by JSPS KAKENHI (Grant Number 18K18072) and a project commissioned by the New Energy and Industrial Technology Development Organization (NEDO).}
\thanks{$^{1}$Mobile Robotics Research Team (MR2T), National Institute of Advanced Industrial Science and Technology (AIST), Ibaraki, Japan
        {\tt\small \{shuji.oishi, k.koide, yokotsuka-masashi, atsuhiko.banno\}@aist.go.jp}}%
}

\begin{document}

\maketitle
\thispagestyle{empty}
\pagestyle{empty}

\begin{abstract}

This study presents a framework for capturing human attention in the spatio-temporal domain using eye-tracking glasses.
Attention mapping is a key technology for human perceptual activity analysis or Human-Robot Interaction (HRI) to support human visual cognition; however, measuring human attention in dynamic environments is challenging owing to the difficulty in localizing the subject and dealing with moving objects.
To address this, we present a comprehensive framework, {\it 4D Attention}, for unified gaze mapping onto static and dynamic objects.
Specifically, we estimate the glasses pose by leveraging a loose coupling of direct visual localization and Inertial Measurement Unit (IMU) values.
Further, by installing reconstruction components into our framework, dynamic objects not captured in the 3D environment map are instantiated based on the input images.
Finally, a scene rendering component synthesizes a first-person view with identification (ID) textures and performs direct 2D-3D gaze association.
Quantitative evaluations showed the effectiveness of our framework.
Additionally, we demonstrated the applications of 4D Attention through experiments in real situations\footnote{See the attached video or \protect\url{https://youtu.be/NG2V3Qs_L8g} for what 4D Attention can offer.}.

\end{abstract}


\section{INTRODUCTION}
\label{sec:introduction}

''{\it The eyes which are the windows of the soul.}`` \\
 \rightline{--- Plato (427 BC - 347 BC)} \\
Eye movements are crucial but implicit cues for determining people's attention.
Gaze estimation enables the study of visual perception mechanisms in humans, and has been used in many fields, such as action recognition\cite{Fathaliyan-Frontiers2018}, situation awareness estimation\cite{Dini-IROS2017}, and driver attention analysis\cite{Maekawa-ICCVW2019}.
It is also a non-verbal communication method, and thus, it can be applied to shared autonomy \cite{Admoni-AAAIS2016} or teleoperation \cite{Webb-ACC2016} in the context of Human-Robot Interaction (HRI).

Recent studies have enabled human attention mapping in 3D environments using mobile eye-tracking glasses\cite{Munn-ETRA2008}\cite{Paletta-IRCV2013}.
Most approaches compute a 3D gaze by extending a measured 2D gaze vector from a camera pose estimated by visual localization or motion capture systems in a pre-built static 3D map.
They are assumed to operate in static environments; however, the real world is a place of constant change, with objects appearing and disappearing from the scenes.
Human attention analysis in both spatial and temporal domains is still an open problem, which when solved will help determine human behavior in the real world.

\begin{figure}[t]
\vspace{2mm}
    \begin{center}
        \includegraphics[width=0.9\linewidth]{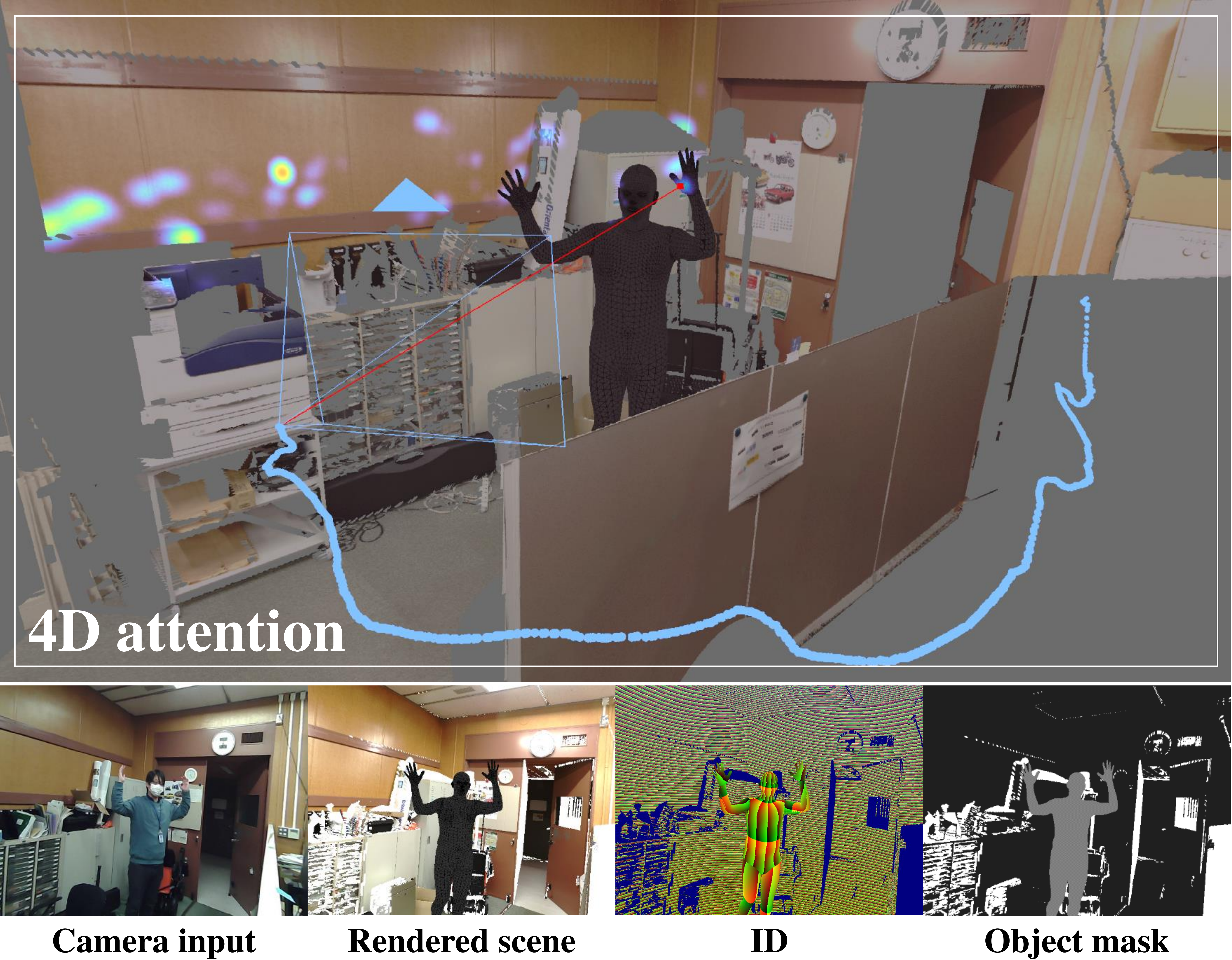}
    \end{center}
    \vspace{-3mm}
    \caption{4D attention analysis in a dynamic environment. Given first-person view with the subject's 2D gaze, it projects human attention onto the static 3D map and dynamic object models employing visual localization, rapid intersection search, and instance object reconstruction.}
    \label{fig:eyecatch}
    \vspace{-5mm}
\end{figure}

To address this issue, we propose a comprehensive framework for 4D attention mapping (see Fig.\ref{fig:eyecatch}).
The main contributions of this study are three-fold:
\begin{itemize}
    \item A new framework, {\it 4D Attention}, is proposed for capturing human attention to static and dynamic objects by assembling 6-DoF camera localization, rapid gaze projection, and instant dynamic object reconstruction. Human attention is accumulated on each 3D mesh model, which makes gaze mapping much more meaningful, for example, the semantic analysis of perceptual activities rather than generating cluttered 3D gaze point clouds. 
    \item The framework is designed so that scene rendering plays a central role. This makes the entire system simple and does not require additional map or object model representations for localization and attention mapping. Additionally, it facilitates a unified attention-mapping procedure regardless of the target objects.
    \item We examined the accuracy and precision of our method using a moving target board whose ground truth position was measured by a total station. Additional experiments for monitoring human attention in the real world demonstrated the capability of analyzing human attention in static and dynamic targets including maps, household items, and people, during the free movement of the subject.
\end{itemize}

\begin{figure*}[t]
\vspace{2mm}
 \begin{center}
    \scriptsize
    \includegraphics[width = 0.9\linewidth ]{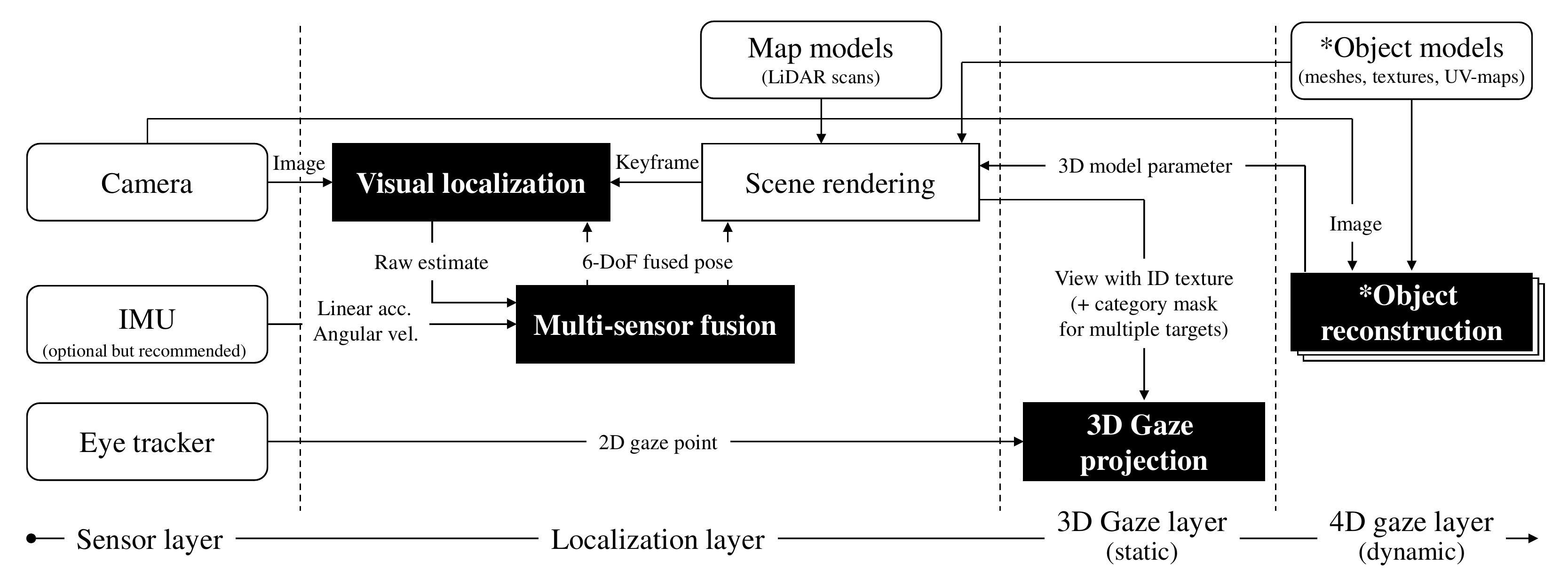}\\
 \end{center}
\vspace{-4mm}
	\caption{Overview of 4D Attention: In the localization layer, we compute the pose of the eye-tracker with C${}^{*}$\cite{Oishi-RAL2020} based on images from a scene camera. Fusing IMU data with the raw pose estimate can further boost and stabilize visual localization\cite{Lynen-IROS2013}. In the 3D gaze layer, the intersection of a gaze ray and the environment map is calculated using the direct 2D-3D gaze association via scene rendering with an ID texture. The 4D gaze layer incorporates any object reconstruction components into the framework to instantiate dynamic objects, which facilitates the analysis of spatio-temporal human attention in the real world. }
\vspace{-4mm}
\label{fig:overview}
\end{figure*}

\section{RELATED WORK}
\label{sec:related_work}

{\bf Eye movement patterns: }
Eye movements imply visual perception activities.
Several approaches have inferred or determined perceptual activities based on the observations from electrooculography (EOG).
Bulling \etal \cite{Bulling-TPAMI2011} developed a pioneering system that classifies several activities from eye movement patterns by utilizing machine learning.
Ishimaru \etal \cite{Ishimaru-UbiComp2014} also determined daily activities including typing, reading, eating, and talking, using signals from EOG glasses.
This approach allows us to identify the current activity of a subject without complex settings, and can be applied to HCI to provide relevant services.

{\bf 2D contextual analysis: }
However, human beings live in a context.  
Visual perception activities are not independent of the surrounding environment; in fact, they are induced by ``attractive'' objects in the scene.
Eye-tracking and gaze overlay on 2D camera views make it possible to determine the focus of the subject, as in \cite{Pelz-EI2011}.
For semantic human attention analysis in natural environments, Fritz and Paletta \cite{Fritz-ICIP2010} introduced object recognition in mobile eye tracking using local image descriptors.
A similar approach can be observed in \cite{Toyama-ETRA2012}, which identifies objects fixated by the subject for a museum guide.
\cite{Harmening-SAGA2013} further progressed toward online object-of-interest recognition using a hierarchical visual feature representation.

{\bf 3D gaze mapping: }
For the holistic estimation of human attention, recent techniques have attempted to obtain fixations in the real 3D world leaving the image plane.
\cite{Pfeiffer-ETRA2012} and \cite{Dini-IROS2017} extended 2D gaze mapping by combining it with a motion capture system to track the pose of gaze glasses, which enables the measurement of the 3D point of interest. 
\cite{Pfeiffer-ETRA2016a} built a similar system relying on visual markers for monocular camera tracking and 3D gaze analysis.
However, they require a complex setup of multiple sensors, making the measurement area small and unscalable to large environments.
Thus, several approaches compute the 3D gaze by localizing an agile monocular camera using visual localization or structure-from-motion.
\cite{Munn-ETRA2008} was the pioneering work, and was followed by promising techniques such as \cite{Paletta-IRCV2013, Hagihara-AH2018} where they estimated camera poses using visual features and projected 3D gaze information onto the pre-built 3D environment map.

{\bf Toward attention analysis in the real world: }
3D gaze mapping facilitates the analysis of human attention regardless of the scale of the environment; however, they still operate only in the static environment.
Attention analysis in dynamic situations is still an open problem; it is necessary to address the {\it spatio-temporal} attention analysis to truly comprehend perceptual activities in the real world.

\section{PROPOSED METHOD}
\label{sec:proposed_method}

\subsection{System overview}

In this study, we propose a comprehensive framework to capture {\it 4D human attention}, which is attention in the spatial and temporal domains in dynamic environments.
A schematic overview of the proposed system is depicted in Fig.\ref{fig:overview}.
Obtaining 4D human attention from eye-tracking glasses with a scene camera has three main problems that need to be solved: robust camera localization, rapid 3D gaze mapping, and instant processing of dynamic objects.

Principally, 4D attention mapping is performed by projecting a first-person 2D human gaze onto a 3D environment map (static) and moving objects (dynamic).
It first requires accurate and stable 6-DoF camera localization even in dynamic environments, which means that appearance of the pre-built 3D map and current view can be significantly changed.
Additionally, given the camera pose, the system has to compute the intersection of the gaze ray and target object surface in real-time to record the 3D distribution of the subject's interest.
Furthermore, dynamic objects such as humans or daily objects should not stay in the same position, but should rather change their poses.
Therefore, they cannot be captured in the 3D map in advance; instead, they should be processed on the fly.

In this section, we describe the major components of the framework shown in Fig.\ref{fig:overview} that are assembled to address these issues and capture 4D attention in the real world.

\begin{figure*}[t]
\vspace{2mm}
 \begin{center}
    \scriptsize
    \includegraphics[width = 0.9\linewidth]{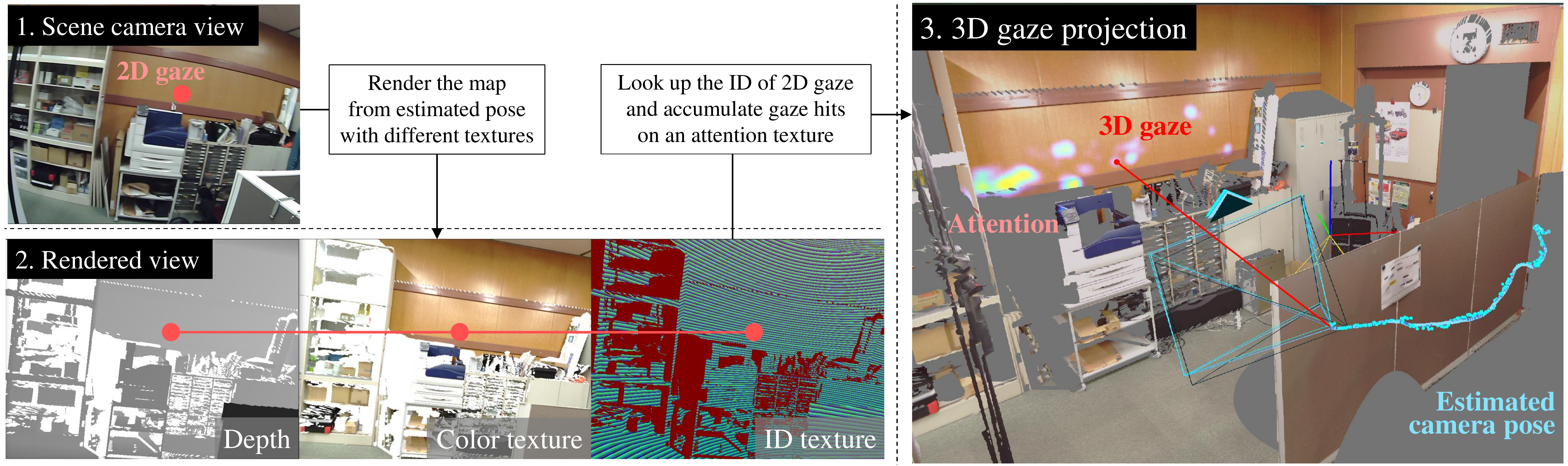}\\
 \end{center}
\vspace{-3mm}
	\caption{ID texture mapping for direct 2D and 3D gaze association: The scene rendering component synthesizes an image from the estimated camera pose to simulate the first-person view. Different textures help to comprehend the scene. Here, we attach an ID texture to the 3D environment map to directly look up the surface with which the input 2D gaze intersects. Gaze hits are accumulated on an attention texture \cite{Pfeiffer-ETRA2016b}, which simplifies the management of human attention information.}
\vspace{-2mm}
\label{fig:gaze_mapping}
\end{figure*}

\subsection{Localization}
\label{subsec:localization}

\subsubsection{Monocular camera localization}

Visual localization is used to infer the pose of an agile monocular camera in a given 3D map.
It can be categorized as either indirect methods via feature point matching, or direct methods via appearance comparison.
Although major 3D gaze mapping methods\cite{Paletta-IRCV2013}\cite{Hagihara-AH2018} rely on indirect methods to estimate the camera pose, they require the construction and maintenance of an extra feature point 3D map for localization.
As will be explained later in Section \ref{subsec:gaze_mapping}, the subject's gaze is projected and accumulated on the dense 3D environment map (or dynamic object models); thus, the requirement doubles the map building cost.
It also incurs other problems such as a 7-DoF exact alignment (including scale) between the environment and feature point maps.

Therefore, for a simple and straightforward system, we employ a direct localization method, specifically C${}^{*}$\cite{Oishi-RAL2020}, which facilitates the localization of the agile monocular camera only with the colored 3D environment map.
It leverages the information-theoretic cost, the Normalized Information Distance (NID), to directly evaluate the appearance similarity between the current camera view and 3D map.
It achieves high robustness to large appearance changes owing to lighting conditions, dynamic obstacles, or different sensor properties\cite{Oishi-RAL2020}, and results in minimal effort in map management.

Given the current view $I_t$, C${}^{*}$ estimates the camera pose $\mathbf{T}_W^t$ in the world coordinate system $W$ via $\mathbb{SE}(3)$ local tracking against a synthetic key frame $I_k$ rendered at a known pose $\mathbf{T}_W^k$:
\begin{equation}
    \begin{aligned}
	    \hat{\mathbf{T}}_k^t & = {\argmin_{\mathbf{T}_k^t}} \; \; {\delta I}_{NID} \left(I_t, I_k, \mathbf{T}_k^t \right), \\
	    \mathbf{T}_W^t & = \hat{\mathbf{T}}_k^t \circ \mathbf{T}_W^k.
    \end{aligned}
\label{eq:minimization}
\end{equation}
C${}^{*}$ reduces the localization problem to alternate local tracking and occasional key frame rendering for efficiency, which leads to 6-DoF real-time localization regardless of the 3D map scale.

The NID metric between the current frame $I_t$ and key frame $I_k$ is given as follows: 
\begin{equation}
	{\delta I}_{NID}\left(I_t, I_k, \mathbf{T}_k^t \right) \equiv \frac{H_{t,k}\left(\mathbf{T}_k^t \right)-I_{t,k}\left(\mathbf{T}_k^t \right)}{H_{t,k}\left(\mathbf{T}_k^t \right)}
	\label{eq:NID}
\end{equation}
where $H_{t,k}$ and $I_{t,k}$ denote the joint entropy and mutual information calculated based on the color co-occurrence in $I_t$ and $I_k$, respectively.
To determine the most likely relative pose $\mathbf{T}_k^t$, gradient-based optimization is performed. 
Specifically, starting from the given initial guess or previously estimated pose, the BFGS is employed to iteratively solve Eq.\ref{eq:minimization} according to the Jacobian of the NID as follows:
\begin{equation}
    \begin{aligned}
	    {}^{(i+1)} \mathbf{T}_k^t & = {}^{(i)} \mathbf{T}_k^t - \alpha B_k^{-1} \frac{d \delta I_{NID}}{d {}^{(i)}\mathbf{T}_{k}^{t}}, \\
	    \frac{d \delta I_{NID}}{d \mathbf{T}_{k}^{t}} & = \frac{\left(\frac{d H_{t,k}}{d \mathbf{T}_{k}^{t}}-\frac{d I_{t,k}}{d \mathbf{T}_{k}^{t}}\right) H_{t,k}-\left(H_{t,k}-I_{t;k}\right) \frac{d H_{t,k}}{d \mathbf{T}_{k}^{t}}}{H_{t,k}^{2}}.
    \end{aligned}
\label{eq:bfgs}
\end{equation}

\subsubsection{Visual-Inertial integration for rapid head and eye movement tracking}

C${}^{*}$ is capable of providing reliable camera poses at several tens of hertz.
To track the rapid head movements of the subjects, we further fuse the localization results and measurements from an Inertial Measurement Unit (IMU) calibrated to the camera in a loosely coupled manner\cite{Lynen-IROS2013}.
The framework allows us to achieve more than several hundreds of hertz estimation rates according to the IMU rates. 
Simultaneously, it significantly stabilizes visual localization by forming a closed loop that feeds the output pose into the localizer as the next initial guess of the optimization.
Localization boosting and stabilization are suitable for real-time gaze projection, as described in the following section.

\subsection{3D gaze projection onto the environment map}
\label{subsec:gaze_mapping}

Given the camera pose (subject's head pose) and gaze position on the 2D image, the 3D human gaze can be recovered by generating a 3D ray beginning from the camera through the gaze point.
To determine the fixation point, the intersection of the gaze ray and target object must be calculated.

Ray casting can be computationally expensive for real-time operation.
Therefore, Paletta \etal \cite{Paletta-IRCV2013} pre-computed a hierarchical map representation, specifically, an Oriented Bounding Box Tree (OBB-Tree), and traversed the tree to rapidly find the intersection.
In \cite{Takemura-ToHMS2014} and \cite{Matsumoto-MobileHCI2019}, the authors estimated the 3D gaze point by first applying Delaunay triangulation to the feature point map, detecting the triangular plane that includes the 2D gaze, and finally investing the sub-mesh 3D gaze point into the world coordinate system from the triangle vertices.
Although these methods work efficiently, they require pre-computation to build certain data structures for 3D gaze mapping, and their resolutions significantly affect the balance between the runtime computation cost and mapping accuracy.
Furthermore, when dealing with dynamic objects that are not included in the pre-built 3D environment map, a more flexible scheme that does not require the construction of the data structure each time is preferable.

Thus, for a unified framework of human gaze projection, we propose ID texture mapping as depicted in Fig.\ref{fig:gaze_mapping}.
Texture mapping is a very popular method for attaching a highly detailed appearance to a geometric model that provides realistic rendering images.
Given a 3D mesh model, its texture image, and per-vertex UV coordinates, we can generate a textured 3D model with GPU acceleration.
Any texture images are available in texture mapping; therefore, we attach a 32-bit integer texture that contains an unique ID of each pixel in its position, for example, $p(x, y) = y * width + x$, for gaze projection.
Specifically, we determine the pixels that are currently observable by rendering the 3D map from the camera pose with the ID texture, and directly find the 3D gaze point by accessing the pixel corresponding to the 2D gaze point.

In addition to the simple setup and direct 2D-3D gaze association, the framework offers other benefits with the use of different types of textures.
For example, by preparing another texture filled with zero and counting gaze hits, attention accumulation can be easily managed on a 2D image similar to the {\it attention texture} proposed in \cite{Pfeiffer-ETRA2016b}.
Additionally, overlaying a texture with an object class or semantics on the ID texture enables the semantic understanding of the subject's perceptual activities \cite{Hagihara-AH2018} in a unified pipeline.

ID texture mapping provides a simple yet efficient way of projecting the human gaze onto any geometric model, which is not limited to the map data.
In the next section, we extend this framework to dynamic objects for 4D attention mapping.

\begin{figure}[t]
\vspace{2mm}
\scriptsize
\begin{center}
    \includegraphics[width = 0.9\linewidth]{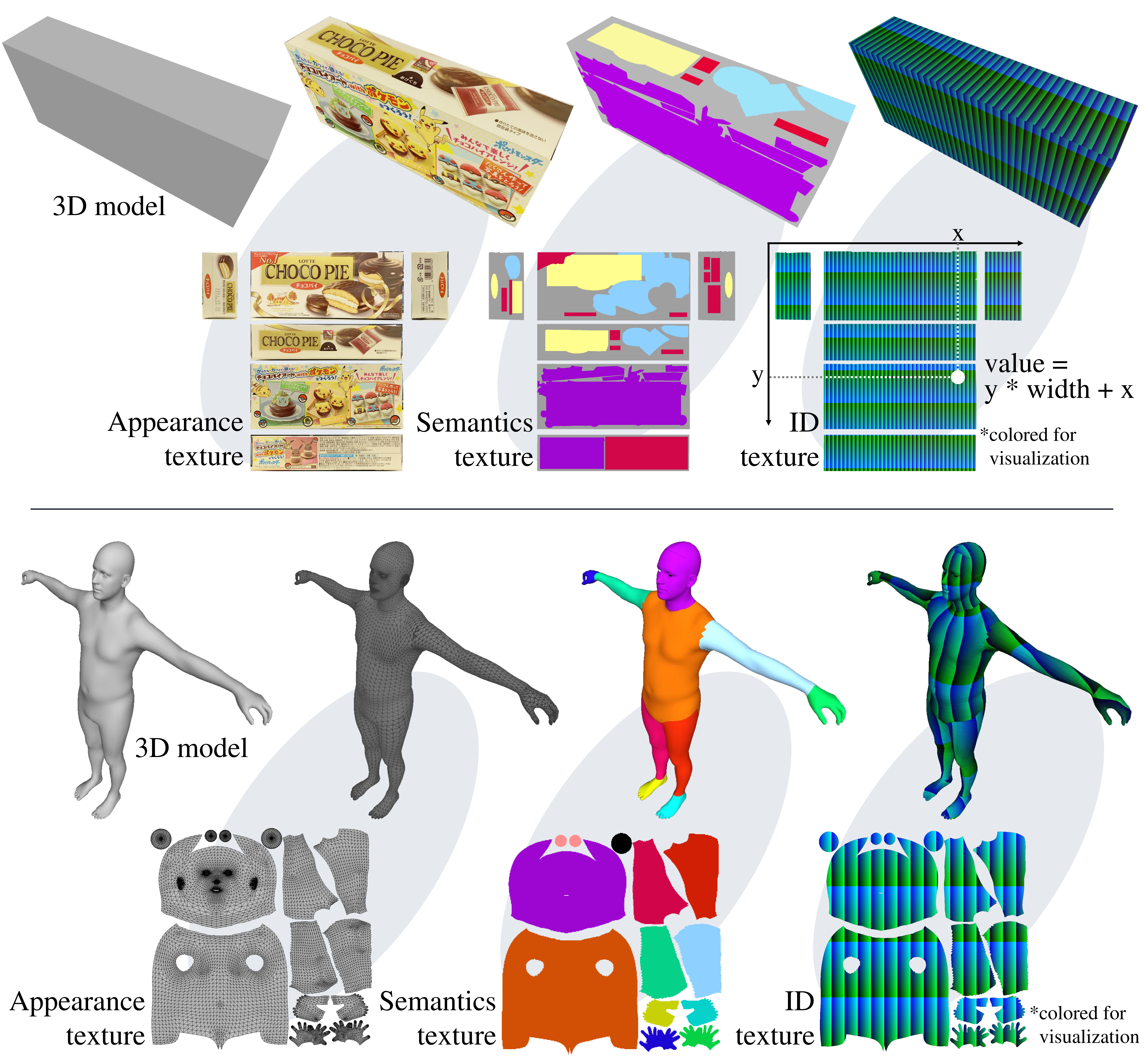}\\
\end{center}
\vspace{-3mm}
\caption{Texturing dynamic objects: Attention mapping onto dynamic objects is performed in the same way as the case of 3D environment maps by employing the corresponding ID textures. Notably, attaching different textures, for example, a semantic texture, helps to determine perceptual activities in the spatio-temporal domain. }
\vspace{-4mm}
\label{fig:dynamic_objects}
\end{figure}

\begin{figure*}[t]
    \vspace{2mm}
    \scriptsize
    \begin{center}
        {\tabcolsep = 2.0mm \begin{tabular}{ccc}
            \includegraphics[height = 0.22\linewidth]{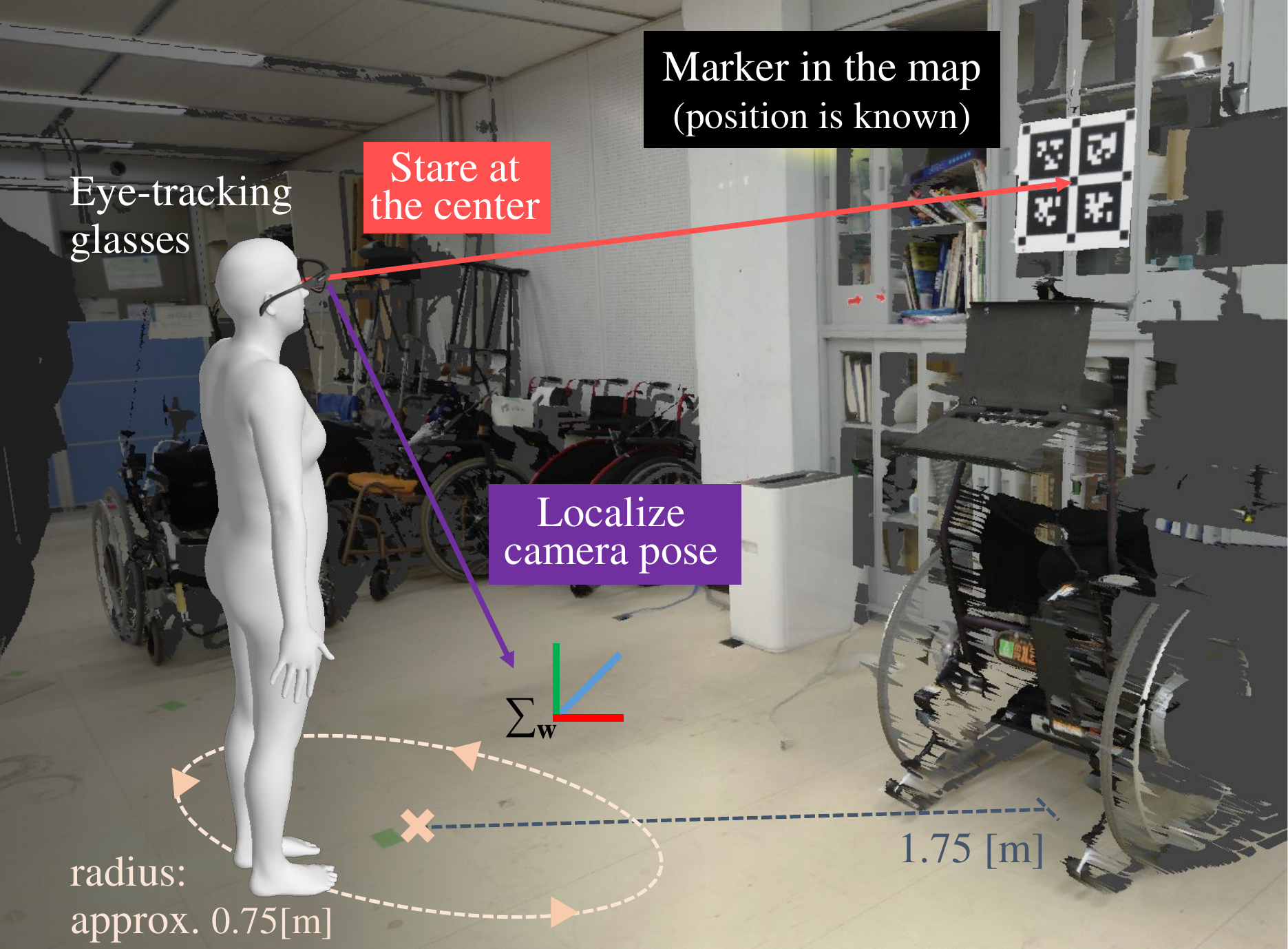} & \includegraphics[height = 0.22\linewidth]{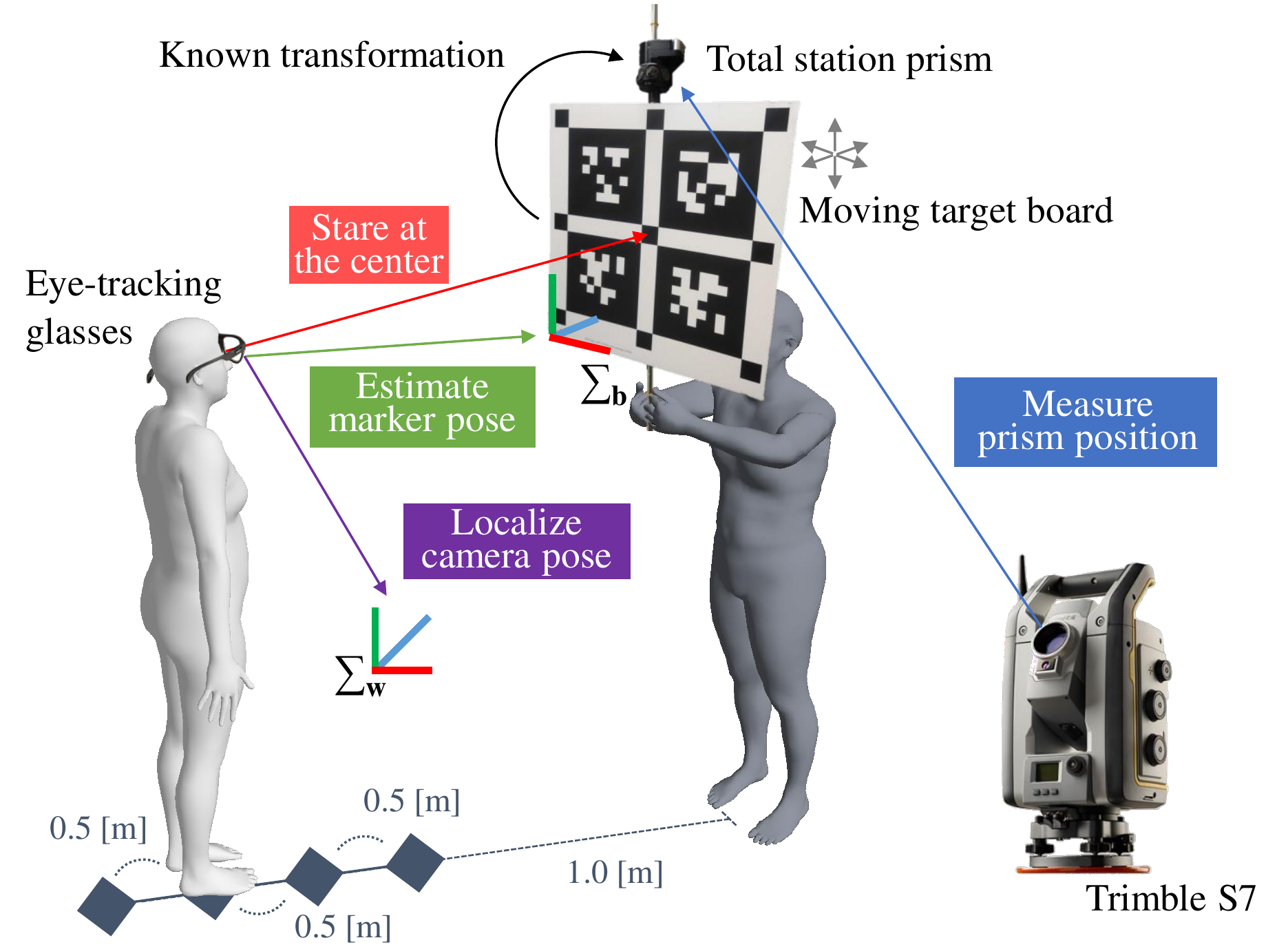} & \includegraphics[height = 0.22\linewidth]{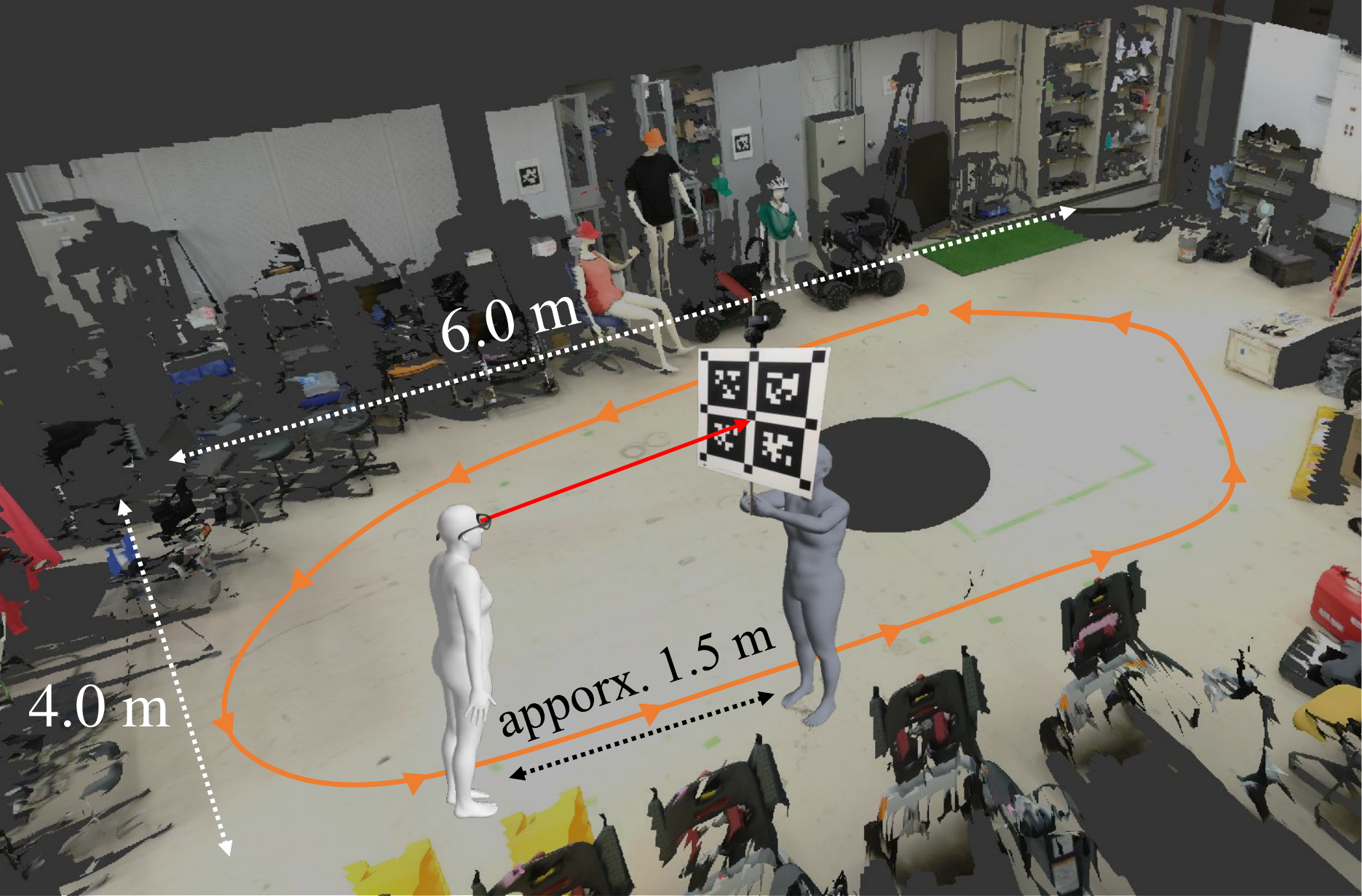} \\
		(a) Evaluation 1: Static target $\times$ Walking around  & (b) Evaluation 2: Dynamic target $\times$ Standing still & (c) Evaluation 3: Dyamic target $\times$ Following \\
        \end{tabular}}
    \end{center}
	\caption{Overview of the quantitative evaluation: AprilTag\cite{Olson_ICRA2011} was used as a target object to evaluate our attention mapping. Our framework generated successive 3D gaze points by finding gaze intersections while the subject stared at its center. The target board and subject changed their states: (Static or dynamic), and (walking around, standing still, or following), respectively, to demonstrate the robustness of the framework to scene dynamics. Notably, AprilTag was embedded in the 3D map in Evaluation 1, whereas it was reconstructed on the fly in Evaluations 2 and 3.}
    \label{fig:quantitative_overview}
    \vspace{-3mm}
\end{figure*}

\subsection{Dynamic object handling for 4D attention mapping}
\label{subsec:dynamic_object_handling}

Objects that do not exist in the map building phase cannot be stored in the 3D environment map, which means that the map data should only record static objects.
However, many dynamic objects such as humans or household items are observed in daily life, and they seem to have ``illegally'' appeared in the static 3D map.
The temporal gap between the mapping and runtime phases causes the absence or presence of dynamic objects, which leads to incorrect gaze projection.

Most conventional works only focus on static scenes and have no choice but to ignore dynamic objects.
To analyze human gaze in dynamic scenes, Fathaliyan \etal \cite{Fathaliyan-Frontiers2018} proposed a 3D gaze tracking method that relies on a marker-based motion capture system installed in a small space.
It inquires the motion capture tabletop objects' poses in a moment and computes the intersections between the object models and gaze vector; however, the settings are costly and the model does not scale to larger environments.
For wearable 3D gaze acquisition outside the laboratory, Qodseya \etal \cite{Qodseya-ECCVW2016} and Hausamann \etal \cite{Hausamann-ETRA2020} developed eye-trackers equipped with depth sensors.
They overlay 2D gaze points on the depth image and directly reconstruct the 3D human gaze.
However, the scheme is highly sensitive to depth noise and the maximum measurement range.
Moreover, the 3D gaze information is represented as cluttered 3D point clouds, which makes gaze analysis less meaningful than accumulation on model surfaces.

To address this, we enable the framework to install additional components of object reconstruction for instantiating dynamic objects not captured in the 3D environment map.
The recent development of object recognition and tracking techniques has facilitated the determination of full 3D shapes of target objects from monocular images on the fly.
Here, we exploit two methods to handle rigid and non-rigid objects, specifically household items and human models, respectively, for 4D attention mapping.
Notably, any desired components that estimate the poses and 3D shapes of specific objects can be incorporated as explained below.

\subsubsection{Household item models (Rigid objects)}

We introduce a pose detection and tracking method\cite{Pauwels-TCSVT2016} into our system.
Given the mesh models and textures of the target objects, it facilitates the recovery of the 6-DoF poses of hundreds of objects in real-time through the proposed scene simulation with SIFT features.
The acquired information is sent to the same process as the 3D environment maps described in Section \ref{subsec:gaze_mapping};
By attaching an ID texture to each model (Fig.\ref{fig:dynamic_objects}) and rendering it at the estimated 6-DoF pose, we can easily associate the 2D human gaze with the object model surface.
Notably, Multiple Render Targets (MRT) on OpenGL are used to create an integer mask image that helps to distinguish the categories and individuals captured in the rendered view (see the bottom right of Fig.\ref{fig:eyecatch}).
In the following experiments, an 8-bit integer mask was rendered in addition to the ID image in the MRT manner to distinguish up to 256 objects belonging to three categories: map, object, and human.

\subsubsection{Human models (Non-rigid objects)}

The human model is a representative example of non-rigid objects that are important for analyzing perceptual activity in the real world.
Humans change their postures unlike rigid objects; therefore, the reconstruction includes non-rigid deformation, making it more complicated than just detecting 6-DoF poses.
In this research, we use the state-of-the-art method, FrankMocap\cite{Rong-arXiv2020}, to instantiate humans in a 3D environment map.
It fits a statistical body model SMPL-X\cite{Pavlakos-CVPR2019} to each person captured in the input image and provides their shape and pose parameters.
The renderer in our framework subscribes the parameters to reconstruct the human models on-demand and examines whether the 3D human gaze hits the surfaces as in the rigid objects.


\section{EXPERIMENTS}
\label{sec:experiments}

\subsection{Setup}

In this section, we verify the capability of the proposed framework to recover 4D human attention in dynamic environments.
We first quantitatively evaluated the accuracy and precision of the recovered gaze points using a dynamic target marker, followed by demonstrations in real situations.

To build 3D environment maps, we used LiDAR, Focus3D (FARO Technologies, Inc.), which enabled us to capture dense and colored 3D point clouds.
A panoramic spherical image can be generated by arranging each vertex color; we used it as a texture of the 3D map while thinning out some vertices to save GPU memory usage.
Notably, our method only assumes that colored or textured 3D models are available for localization and gaze mapping, and thus it also operates on 3D geometric models reconstructed with different sensors, for example, RGB-D SLAM \cite{Lee-CVPR2020}, similar to \cite{Paletta-IRCV2013}.

The rendering and localization components rely on GPU parallelization; a GeForce GTX2080 performed the computations in all the experiments.
We also used a wearable eye tracker, Tobii Pro Glasses 3 (Tobii Technology, Inc.) to capture first-person views with the subject's 2D gaze information and IMU data.

\subsection{Performance evaluation}

To evaluate the proposed attention mapping, AprilTag \cite{Olson_ICRA2011}, which provides reliable 6-DoF marker poses, was employed as shown in Fig.\ref{fig:quantitative_overview}, whereas the subject was changing the relative positions and its states.
We asked the subject to stare at the center of the target board ($0.24 \times 0.24$ [m]) wearing the eye-tracker, and our method generated the corresponding 3D gaze points.
In Evaluation 1, the board was embedded in the 3D map; thus, we calculated the Absolute Position Error (APE) between the generated 3D gaze points and the center of the board.
In Evaluations 2 and 3, the ground truth trajectories of the agile target board were obtained by tracking a total station prism attached to the board with the known relative transformation using a Trimble S7 (Trinble Navigation, Limited.).
Subsequently, we synchronized the pairs of trajectories based on system timestamps to evaluate the Absolute Trajectory Error (ATE)\cite{Zhang-IROS2018} with a least-squares transformation estimation\cite{Umeyama-TPAMI1991}, in addition to APE.
Notably, the 3D trajectory comparison computes a rigid transformation that minimizes the positional errors between the two point clouds.
The minimization process cancels the systematic bias underlying the framework, which is caused by reasons such as eye-camera miscalibration.
Therefore, the ATE is approximately equivalent to the precision of our framework, whereas the APE is equivalent to the accuracy.

\begin{figure}[t]
\vspace{2mm}
  \begin{center}
  \scriptsize
  \includegraphics[width=0.85\linewidth]{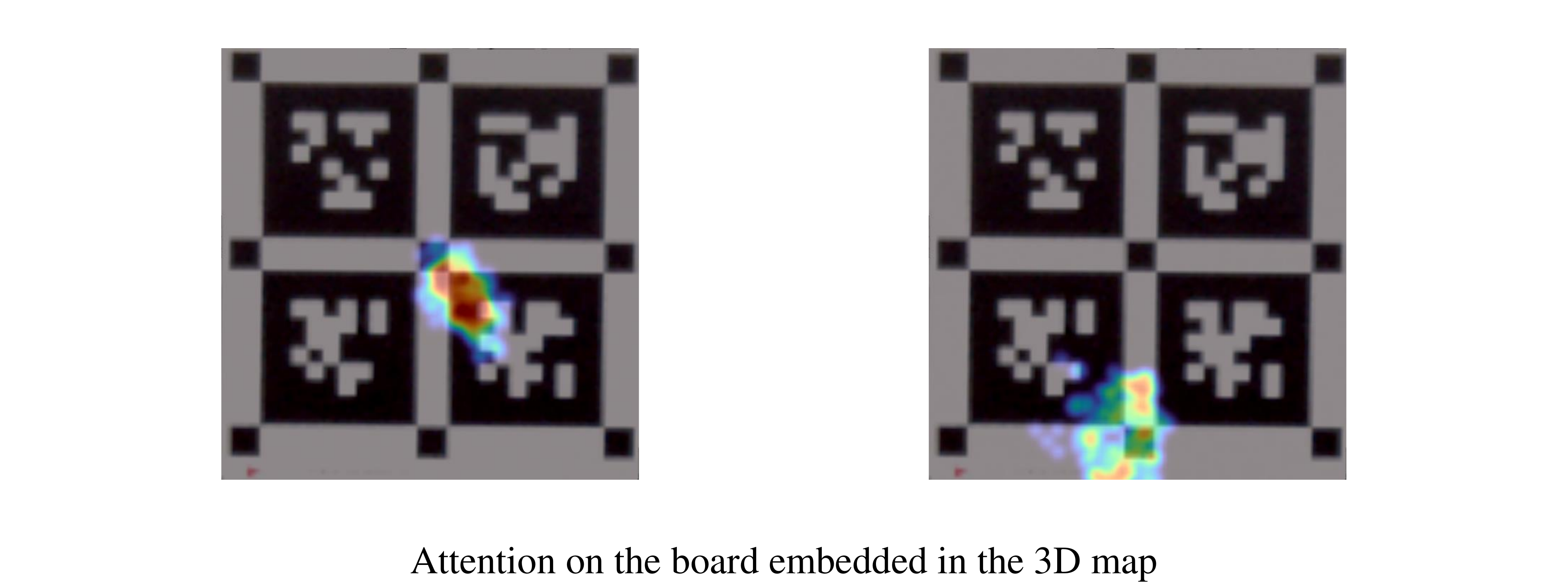} \\
  (a) Evaluation 1: 4D Attention (left) and \cite{Paletta-IRCV2013} (right)\vspace{1mm}\\
  \includegraphics[width=0.85\linewidth]{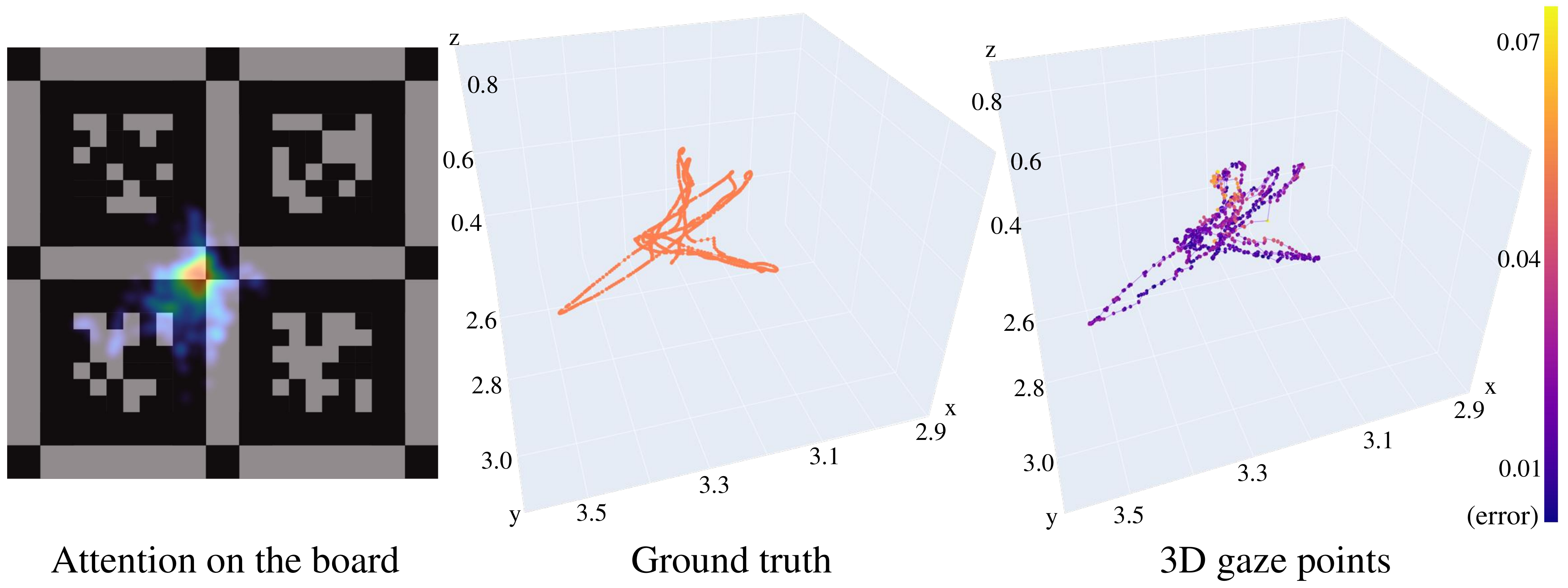} \\
  (b) Evaluation 2 (1 m) \vspace{1mm}\\
  \includegraphics[width=0.85\linewidth]{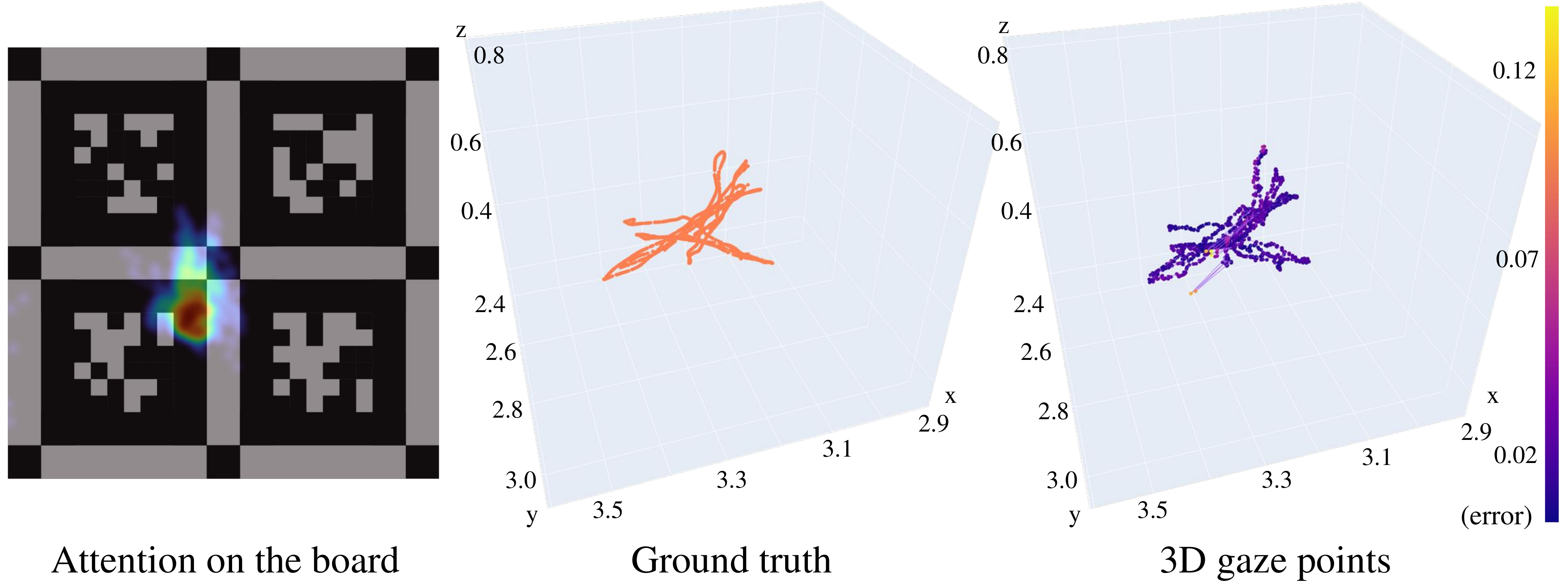} \\
  (c) Evaluation 2 (1.5 m) \vspace{1mm}\\
  \includegraphics[width=0.85\linewidth]{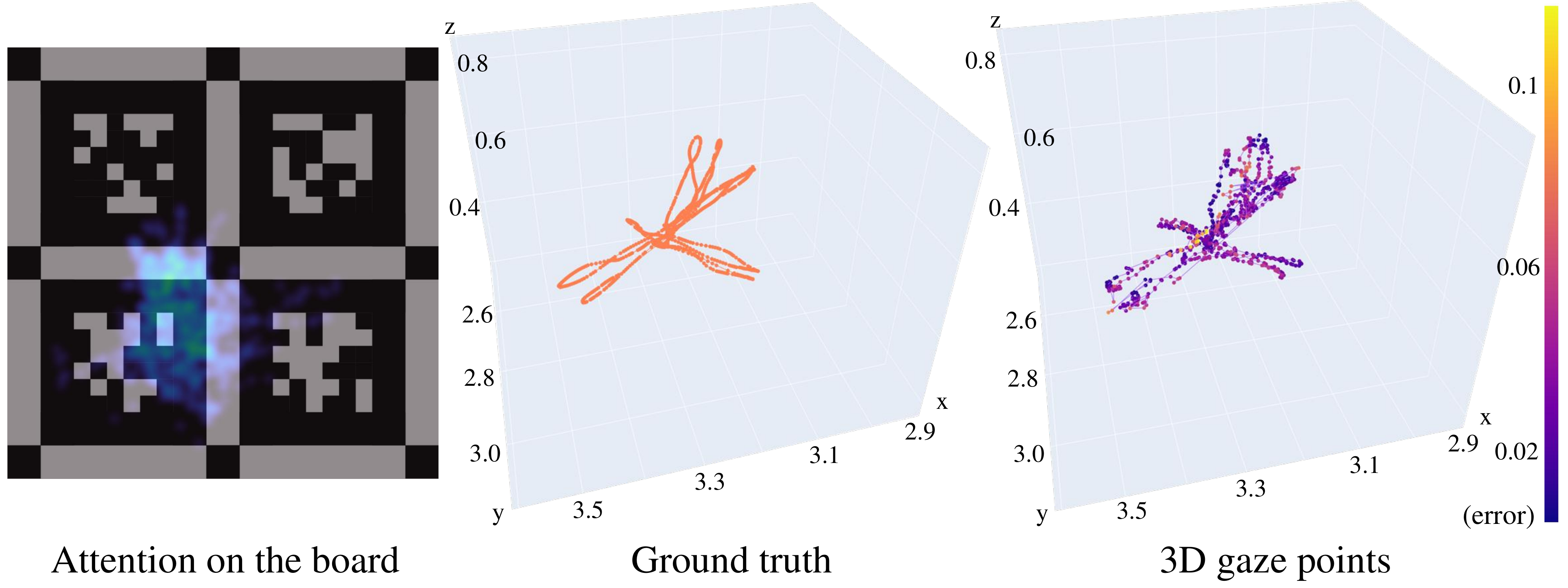} \\
  (d) Evaluation 2 (2 m) \vspace{1mm}\\
  \includegraphics[width=0.85\linewidth]{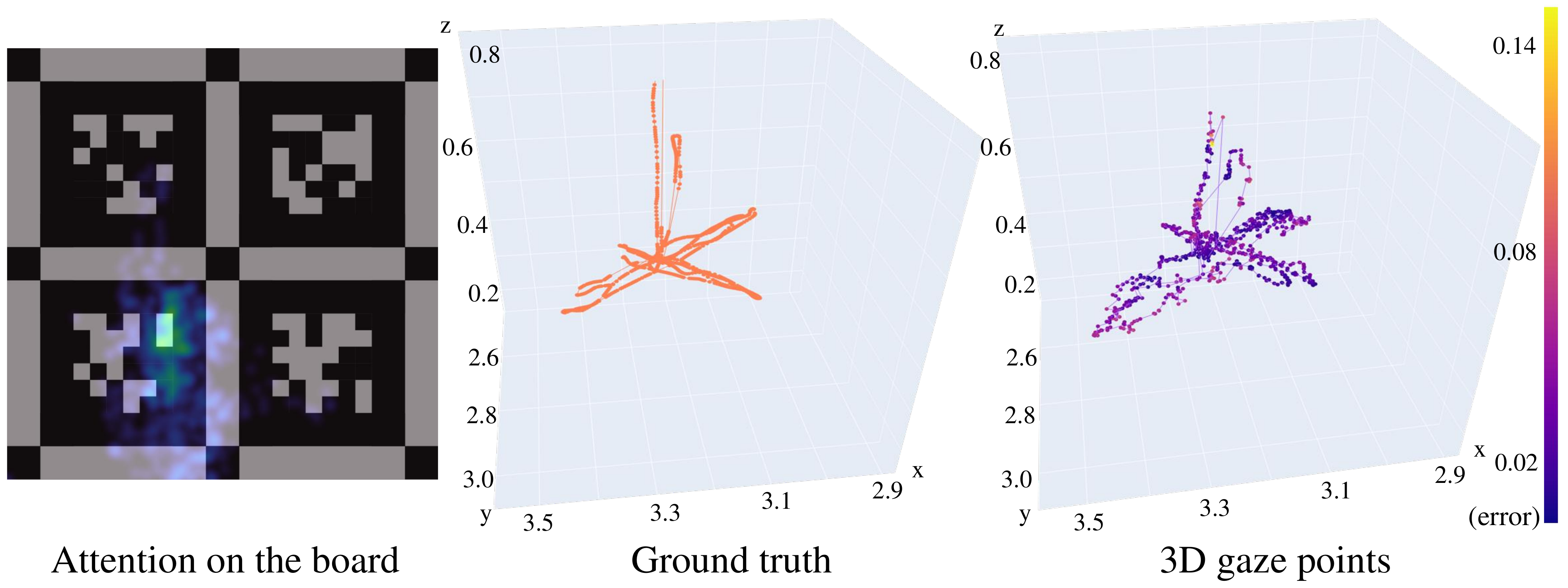} \\
  (e) Evaluation 2 (2.5 m) \vspace{1mm}\\
  \includegraphics[width=0.85\linewidth]{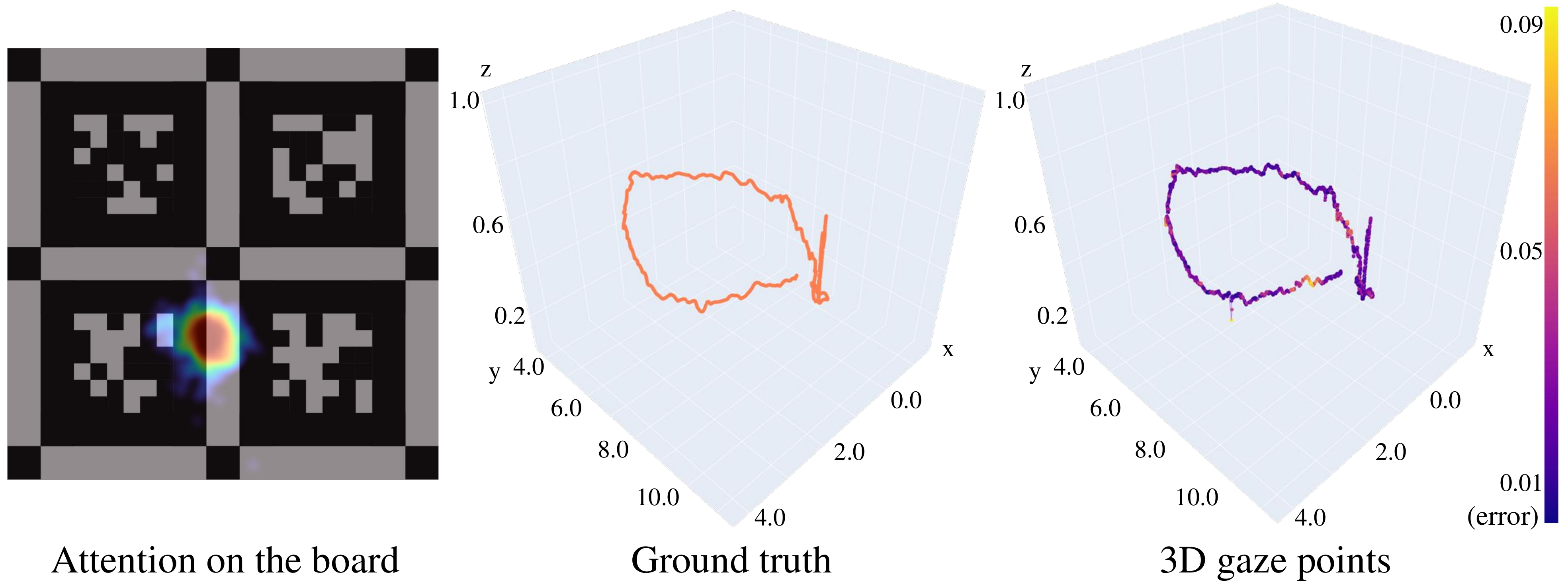} \\
  (f) Evaluation 3 \\
  \end{center} \vspace{-3mm}
	\caption{Quantitative evaluation results: The 3D gaze points obtained in each situation were compared with the ground truth (see also Table \ref{tab:quantitative}). The proposed framework overwhelmed the competitive method\cite{Paletta-IRCV2013} and achieved high-precision 4D gaze mapping in every case. However, the bias was clearly observed in the gaze accumulation, and the accuracy proportionally decreased as the distance from the target board increased. The results imply that our framework is capable of providing stable gaze projection onto dynamic objects, and strict gaze-camera calibration of eye-tracking glasses may improve the accuracy cancelling the systematic error.}
  \label{fig:quantitative}
  \vspace{-4mm}
\end{figure}

Table \ref{tab:quantitative} and Figure \ref{fig:quantitative} present the evaluation results. ---

{\bf Evaluation 1:}
We demonstrated the performance of our framework in a static scene to compare it with the most relevant work \cite{Paletta-IRCV2013} as a baseline.
Specifically, we implemented \cite{Paletta-IRCV2013} whose localizer was replaced with state-of-the-art indirect visual localization\cite{Campos-TRO2021} for a comparison in the same 3D map retaining the concept of the method.
Compared with \cite{Paletta-IRCV2013}, 4D attention achieved high accuracy of 3D gaze mapping benefitting from the rendering-centerd framework such as direct localization and ID texture mapping, which suppress the systematic error.

{\bf Evaluation 2:}
The subject watched the center of the moving target standing at four different positions to evaluate the influence of proximity following the evaluations in previous studies\cite{Paletta-IRCV2013}\cite{ Hagihara-AH2018}.
Overall, although the APE (inaccuracy) increased proportionally with the distance from the target board, the framework successfully suppressed the increase in the ATE (imprecision).

{\bf Evaluation 3:}
The subject walked around a $4 \times 6$ [m] space to follow the moving target board approximately 1.5 [m] behind while watching the center.
Notably, the subject and the person to follow held an assistant rope to maintain their distance.
Although the proposed framework slightly increased the APE and ATE owing to the necessity of the 6-DoF and instant object reconstruction in a complicated situation, it successfully facilitated valid attention mapping even in highly dynamic environments.

\begin{table}[t]
\vspace{2mm}
\caption{Errors of 3D gaze points in the quantitative evaluation.}
  \label{tab:quantitative}
  \scriptsize
  \vspace{-4mm}
  \begin{center}
  {\tabcolsep = 1.3mm \begin{tabular}{l|cc|c|cc} \hline
   No.        & object  & subject     & distance [m] & APE [m]                      & ATE [m]                 \\
      	      & state   &  state      & from board   & ($\simeq$ inaccuracy)        & ($\simeq$ imprecision)  \\ \hline \hline
   1          & static  & walking     & approx.      & $0.034 \pm 0.015$            & -                       \\
              &         & around      & 1.0 - 2.5    & $0.115 \pm 0.021$\textdagger  & -                       \\ \hdashline[0.5pt/1pt]
              &         &             & 1.0          & $0.028 \pm 0.016$            & $0.020 \pm 0.012$       \\
   2          & dynamic & standing    & 1.5          & $0.034 \pm 0.012$            & $0.017 \pm 0.011$       \\
              &         &  still      & 2.0          & $0.049 \pm 0.019$            & $0.034 \pm 0.017$       \\
              &         &             & 2.5          & $0.070 \pm 0.025$            & $0.034 \pm 0.018$       \\ \hdashline[0.5pt/1pt] 
   3          & dynamic & following   & approx. 1.5  & $0.046 \pm 0.0092$           & $0.024 \pm 0.014$       \\ \hline
  \end{tabular}}
  \end{center}
  \vspace{-1mm}
  \textdagger : Errors of 3D gaze points generated by \cite{Paletta-IRCV2013} (our implementation) as a baseline.
  \vspace{-4mm}
\end{table}

 \begin{figure}[t]
\vspace{2mm}
 \begin{center}
  \scriptsize
 \begin{flushleft} first-person view \end{flushleft}
 {\tabcolsep = 0.2mm \begin{tabular}{cccc}
        \includegraphics[width = 0.24\linewidth]{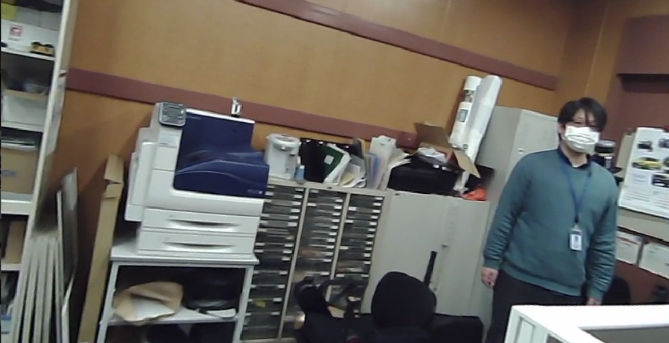} & \includegraphics[width = 0.24\linewidth]{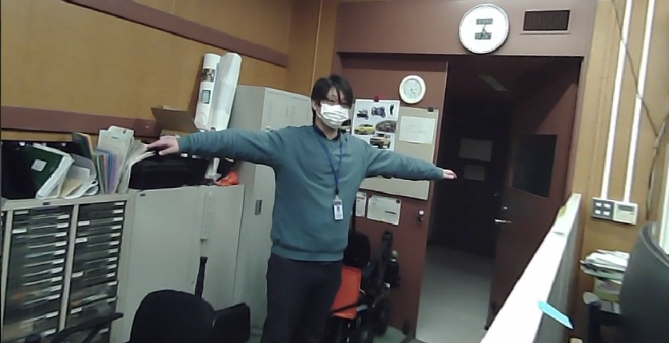} &
        \includegraphics[width = 0.24\linewidth]{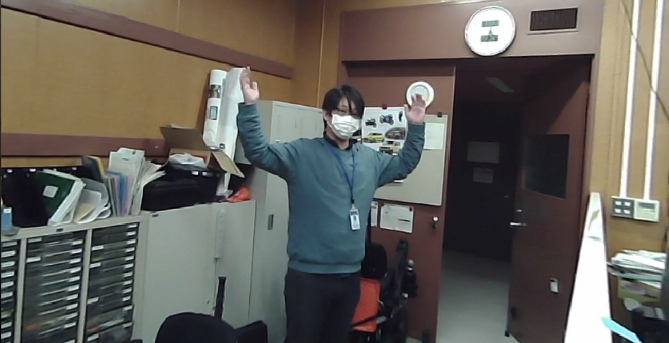} & \includegraphics[width = 0.24\linewidth]{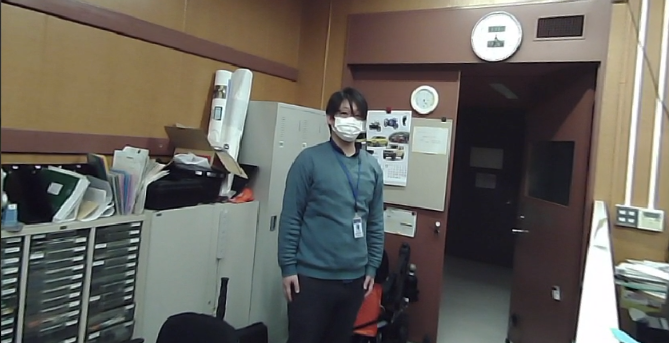} \\
  \end{tabular}} \\ \vspace{-2.5mm}
 \begin{flushleft} 4D attention \end{flushleft} \vspace{-2mm}
 {\tabcolsep = 0.2mm \begin{tabular}{cccc}
	\includegraphics[width = 0.24\linewidth]{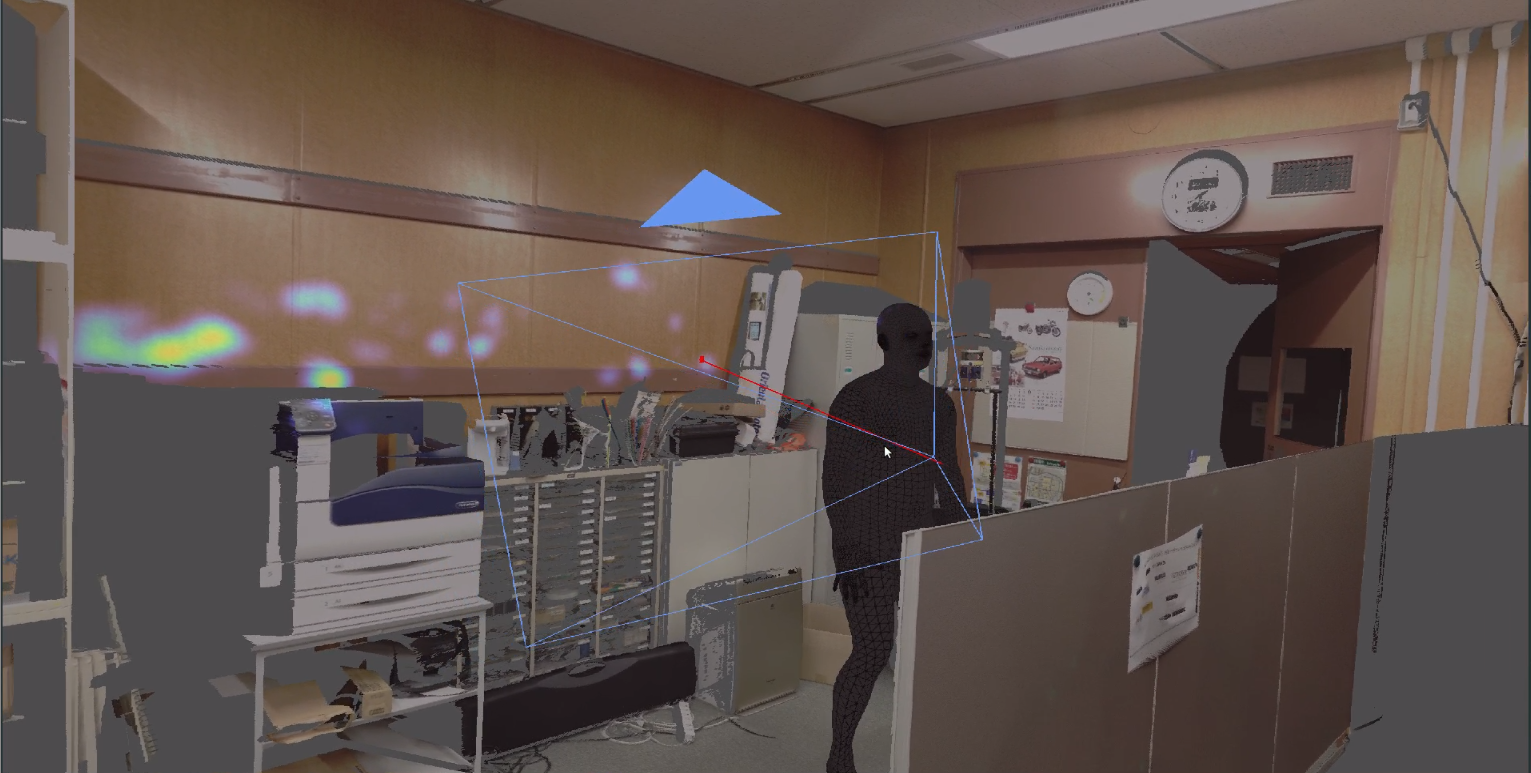} & \includegraphics[width = 0.24\linewidth]{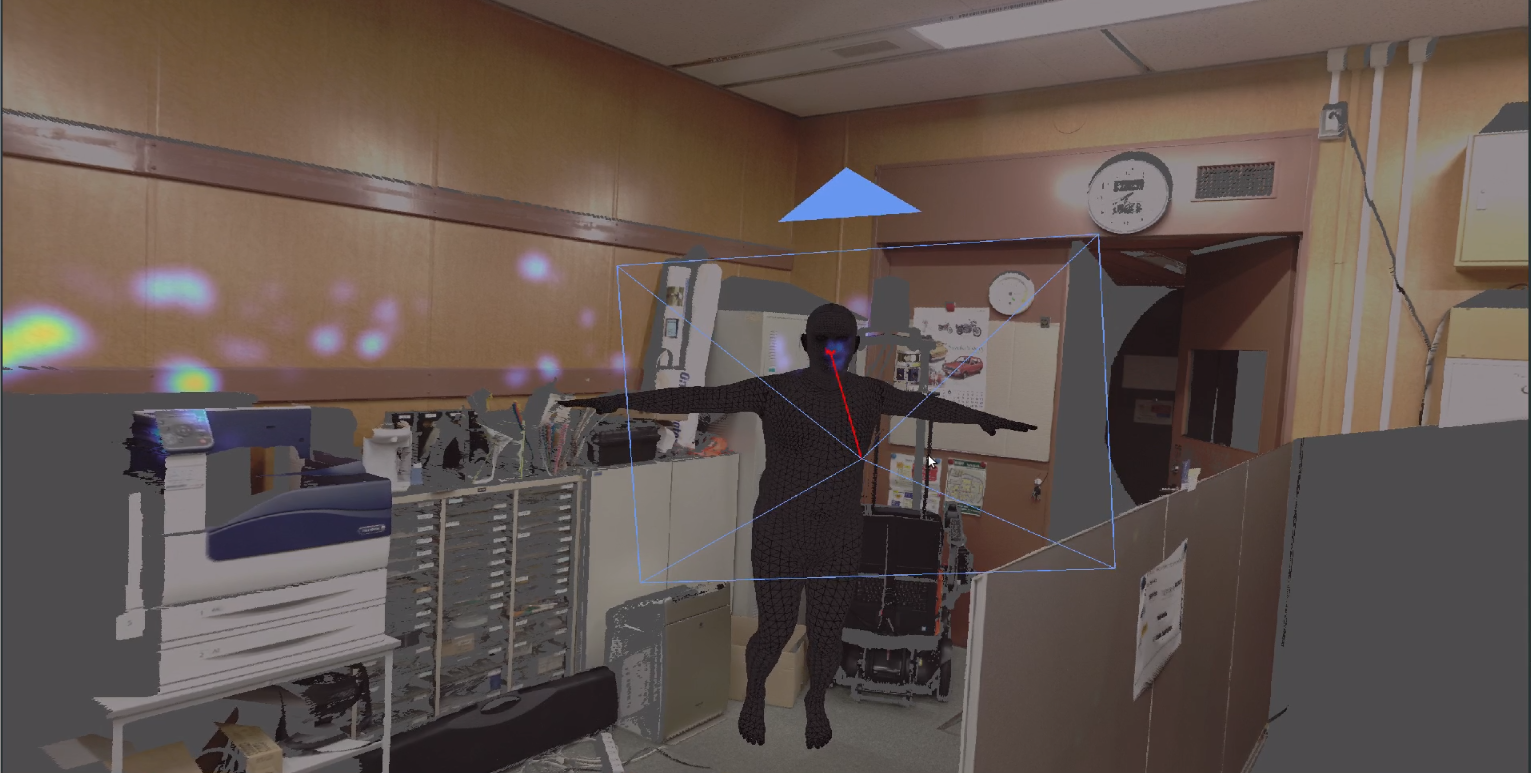} &
	\includegraphics[width = 0.24\linewidth]{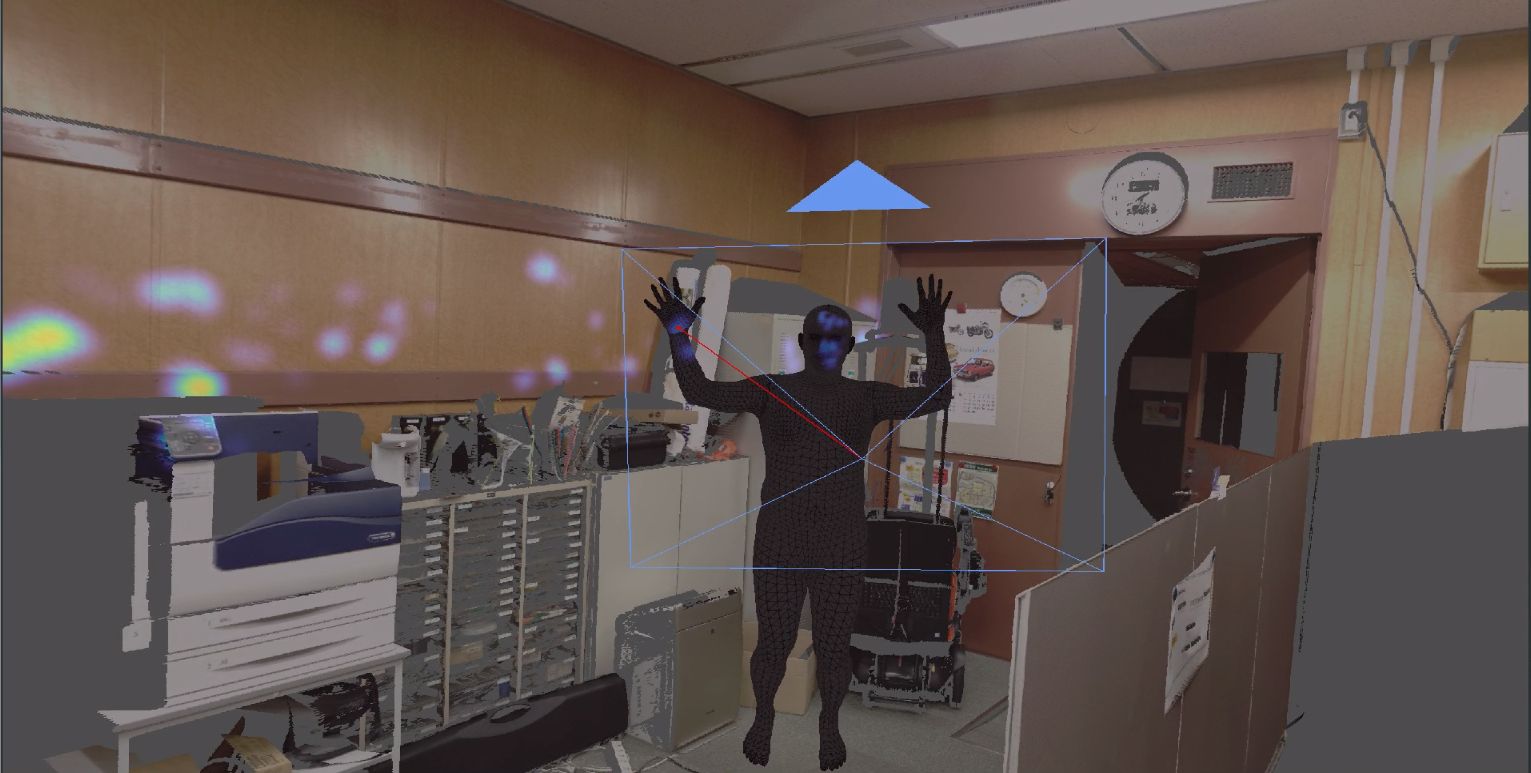} & \includegraphics[width = 0.24\linewidth]{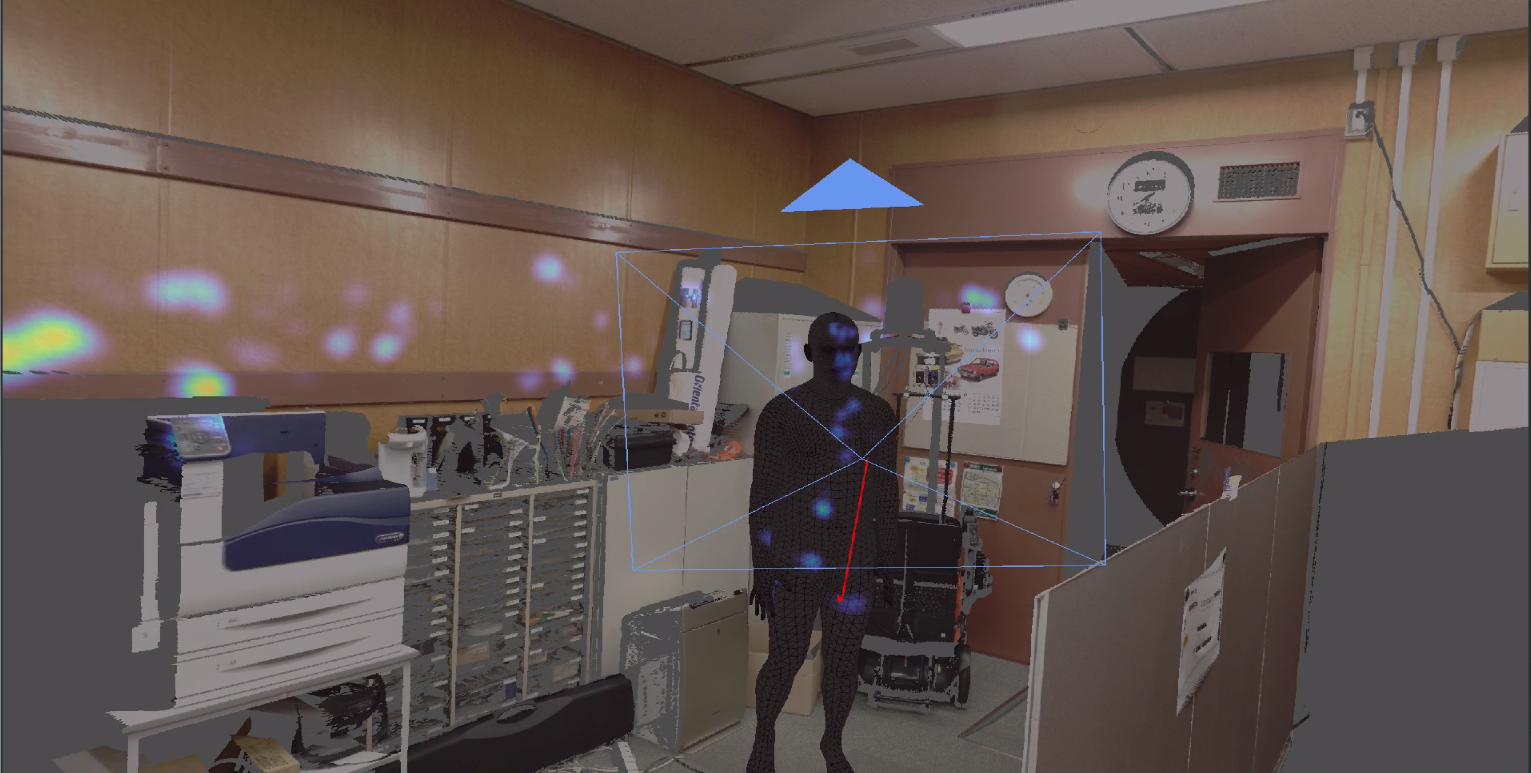} \\
  \end{tabular}}
  \\ (a) Case 1: Observe physical actions of a person \vspace{-1mm}

 \begin{flushleft} first-person view \end{flushleft} \vspace{-2mm}
{\tabcolsep = 0.2mm \begin{tabular}{cccc}
       \includegraphics[width = 0.24\linewidth]{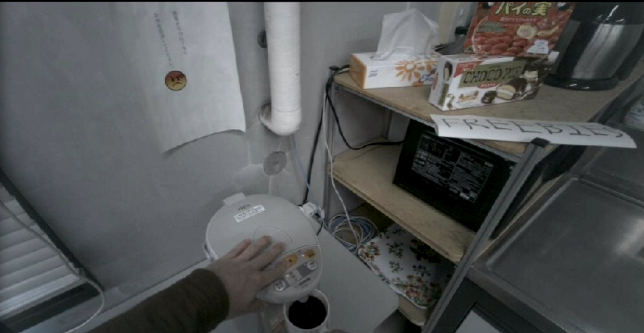} & \includegraphics[width = 0.24\linewidth]{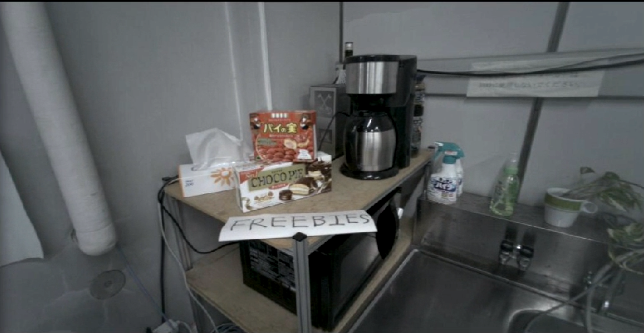} &
       \includegraphics[width = 0.24\linewidth]{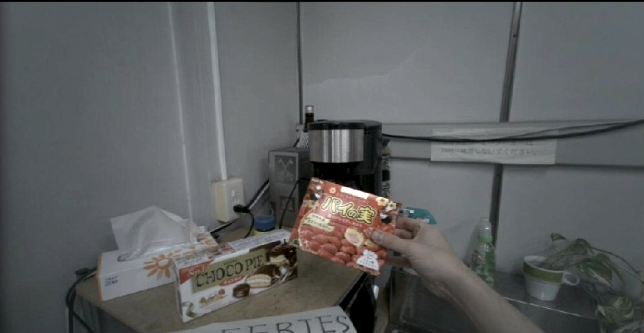} & \includegraphics[width = 0.24\linewidth]{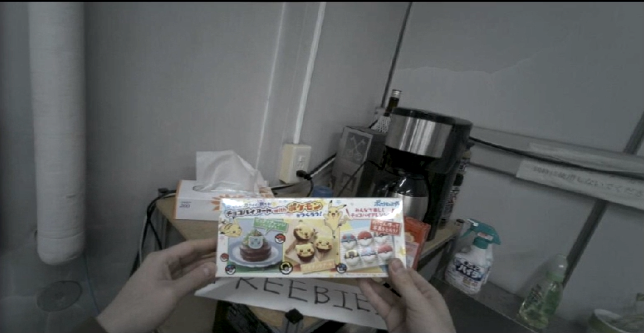} \\
 \end{tabular}} \\ \vspace{-2.5mm}
\begin{flushleft} 4D attention \end{flushleft} \vspace{-2mm}
{\tabcolsep = 0.2mm \begin{tabular}{cccc}
       \includegraphics[width = 0.24\linewidth]{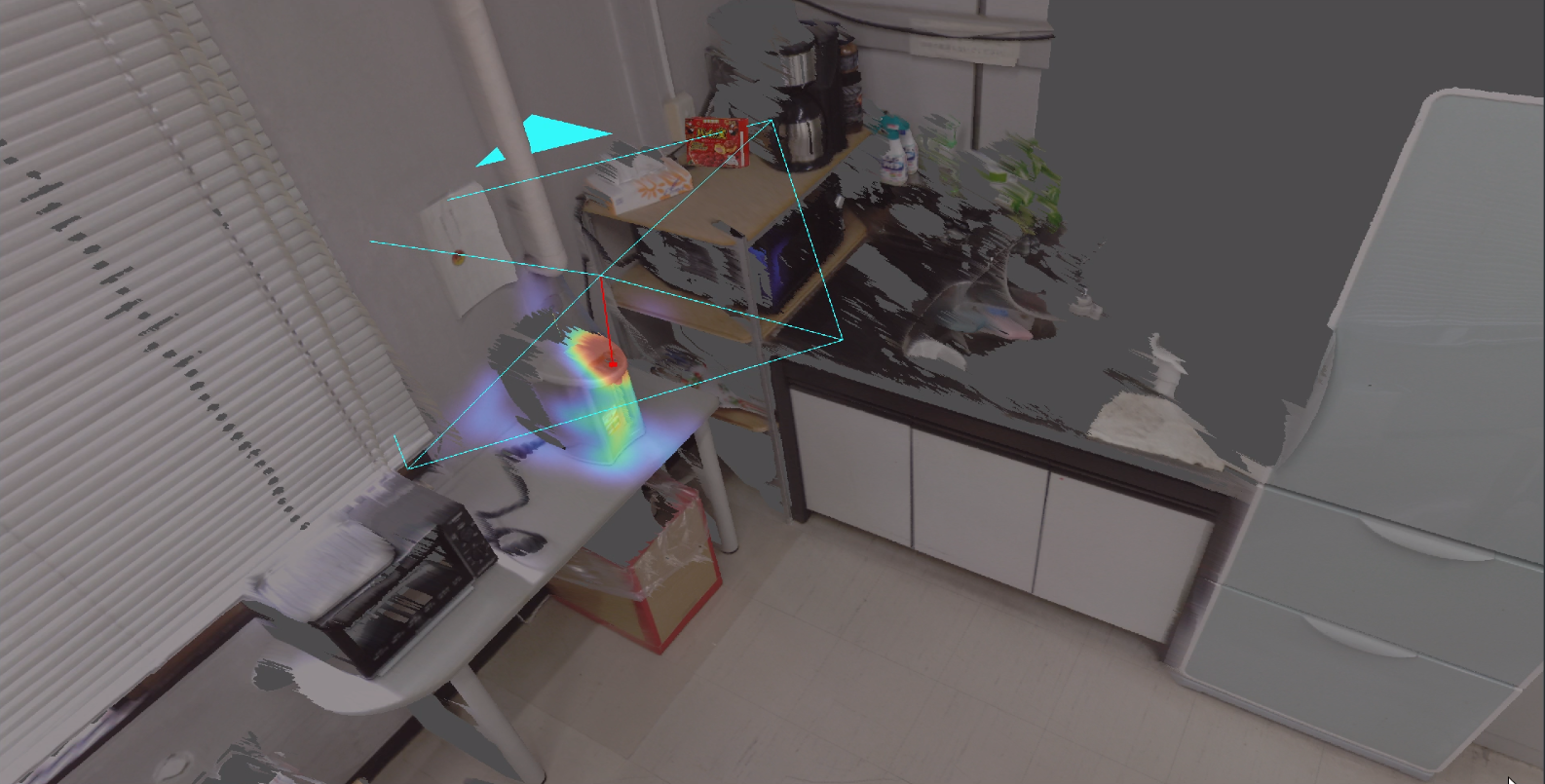} & \includegraphics[width = 0.24\linewidth]{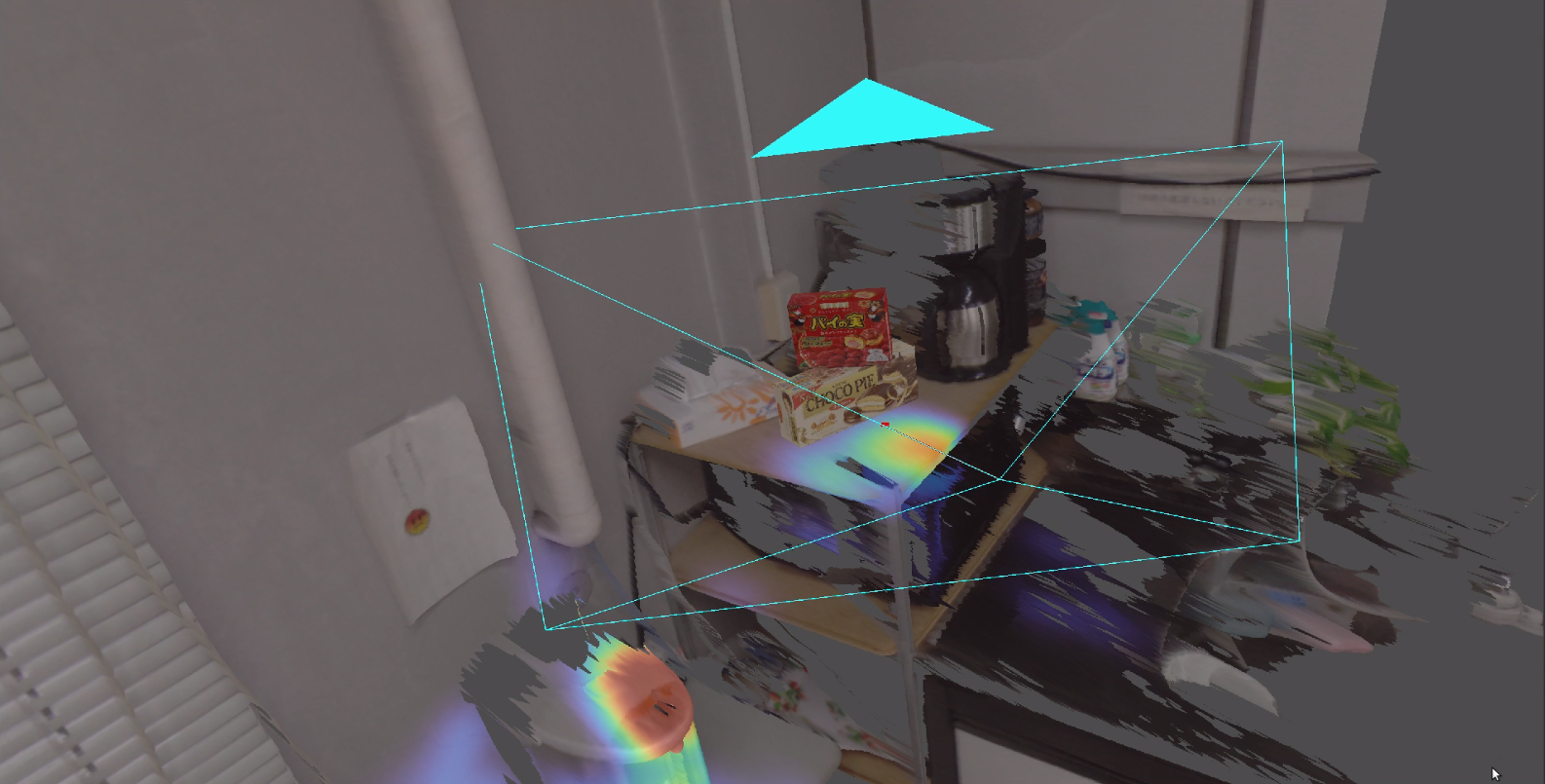} &
       \includegraphics[width = 0.24\linewidth]{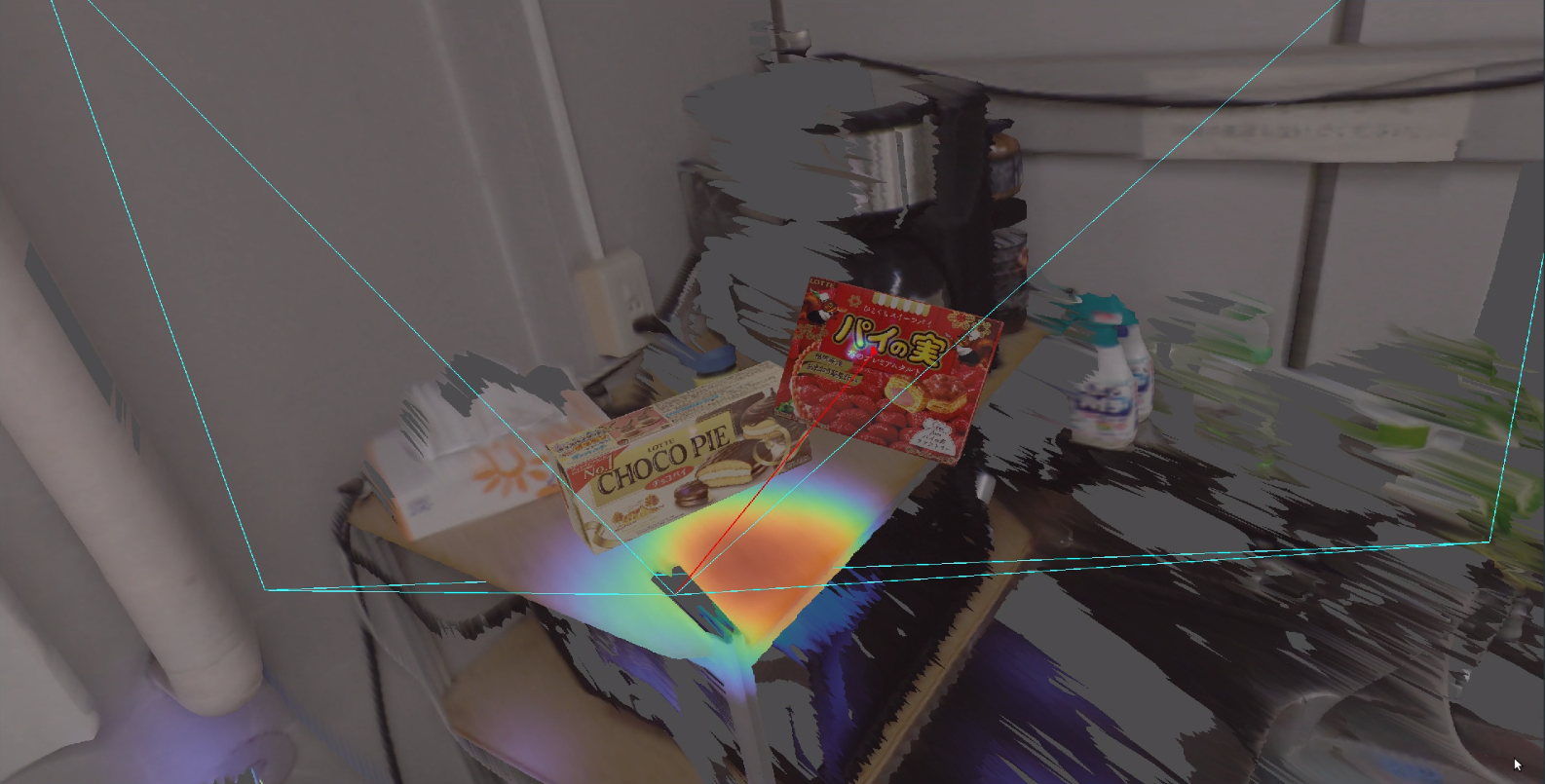} & \includegraphics[width = 0.24\linewidth]{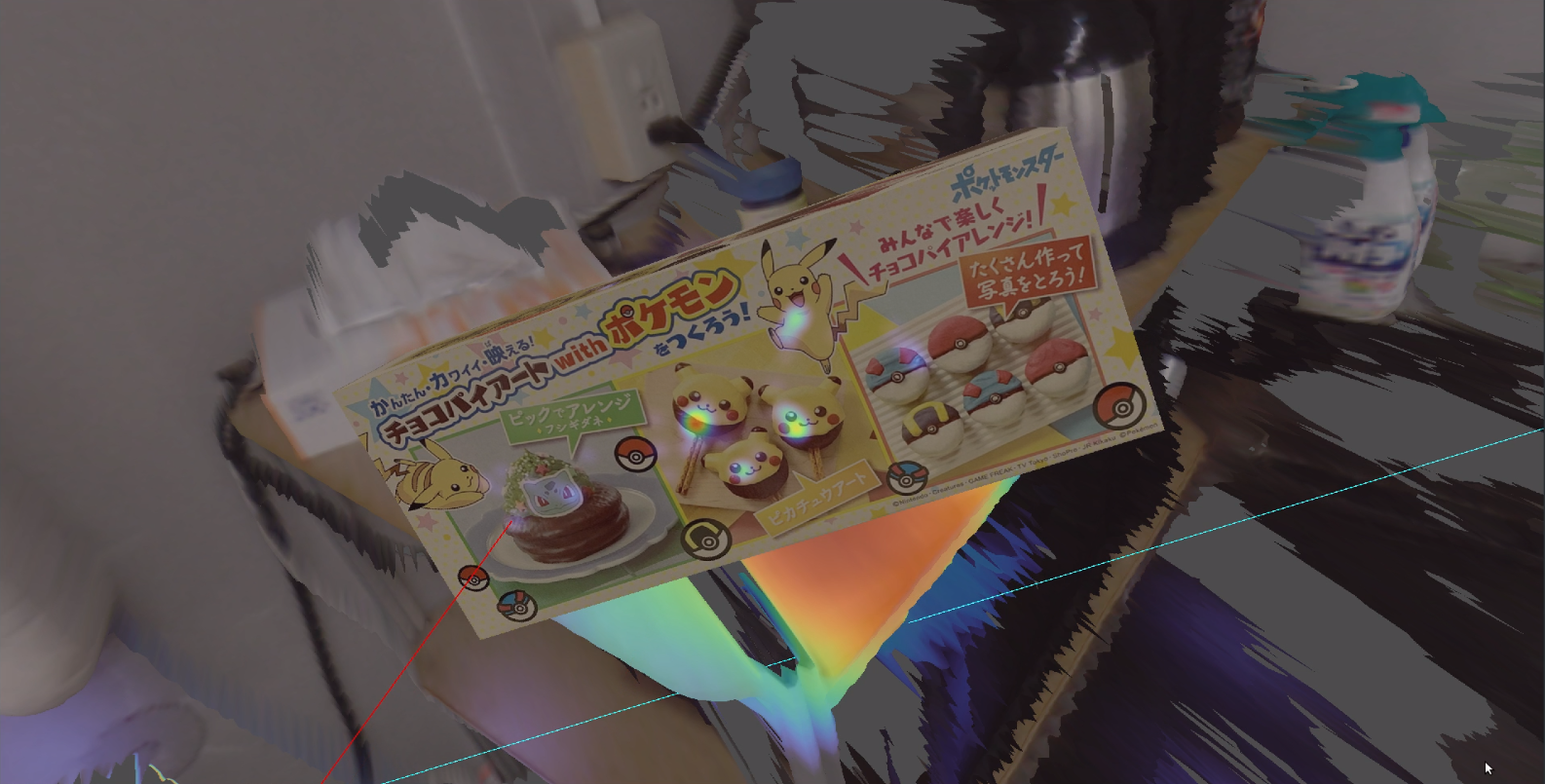} \\
 \end{tabular}}
 \\ (b) Case 2: Take a Coffee break \vspace{-1mm}

\begin{flushleft} first-person view \end{flushleft} \vspace{-2mm}
{\tabcolsep = 0.2mm \begin{tabular}{cccc}
       \includegraphics[width = 0.24\linewidth]{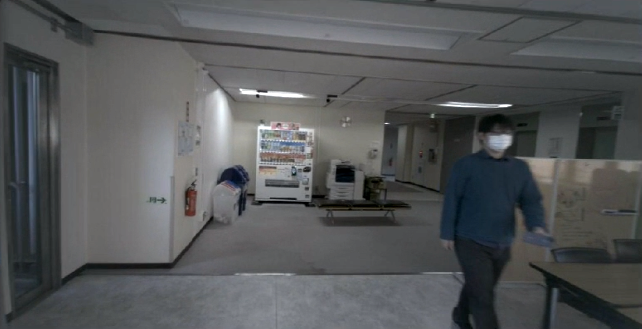} & \includegraphics[width = 0.24\linewidth]{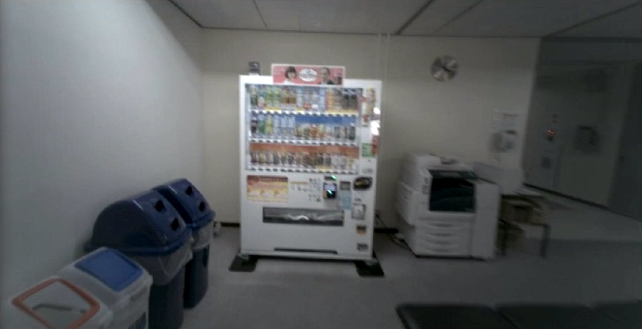} &
       \includegraphics[width = 0.24\linewidth]{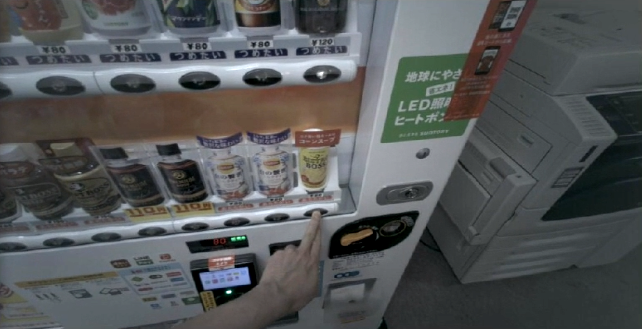} & \includegraphics[width = 0.24\linewidth]{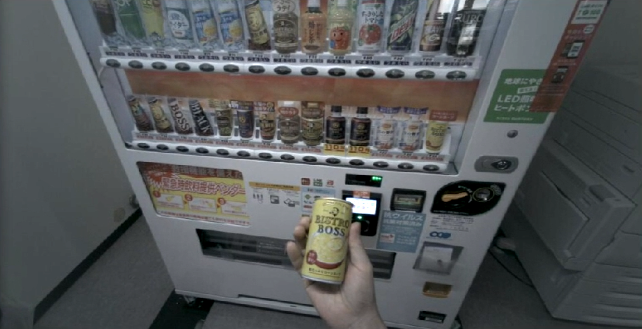} \\
 \end{tabular}} \\ \vspace{-2.5mm}
\begin{flushleft} 4D attention \end{flushleft} \vspace{-2mm}
{\tabcolsep = 0.2mm \begin{tabular}{cccc}
       \includegraphics[width = 0.24\linewidth]{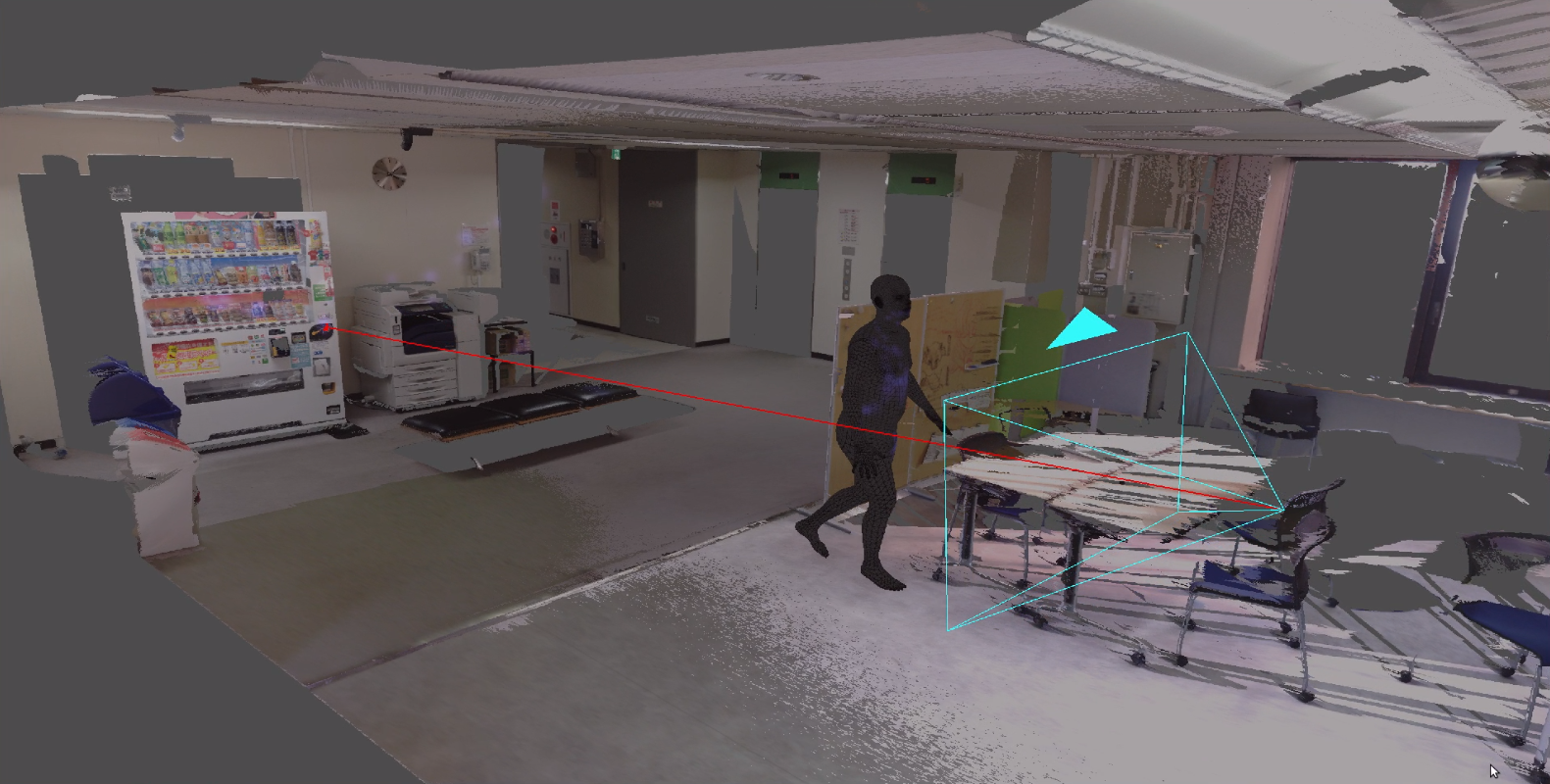} & \includegraphics[width = 0.24\linewidth]{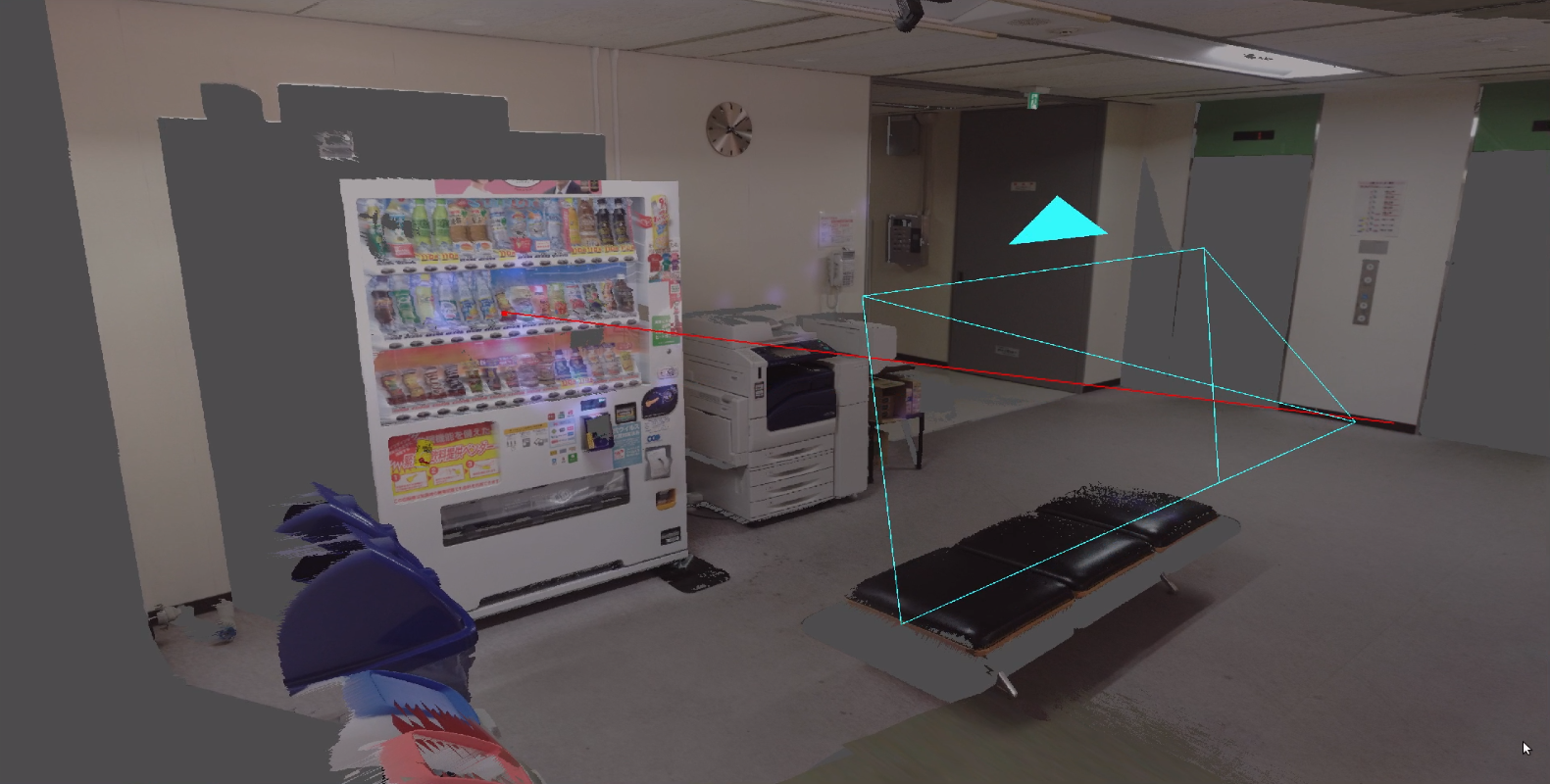} &
       \includegraphics[width = 0.24\linewidth]{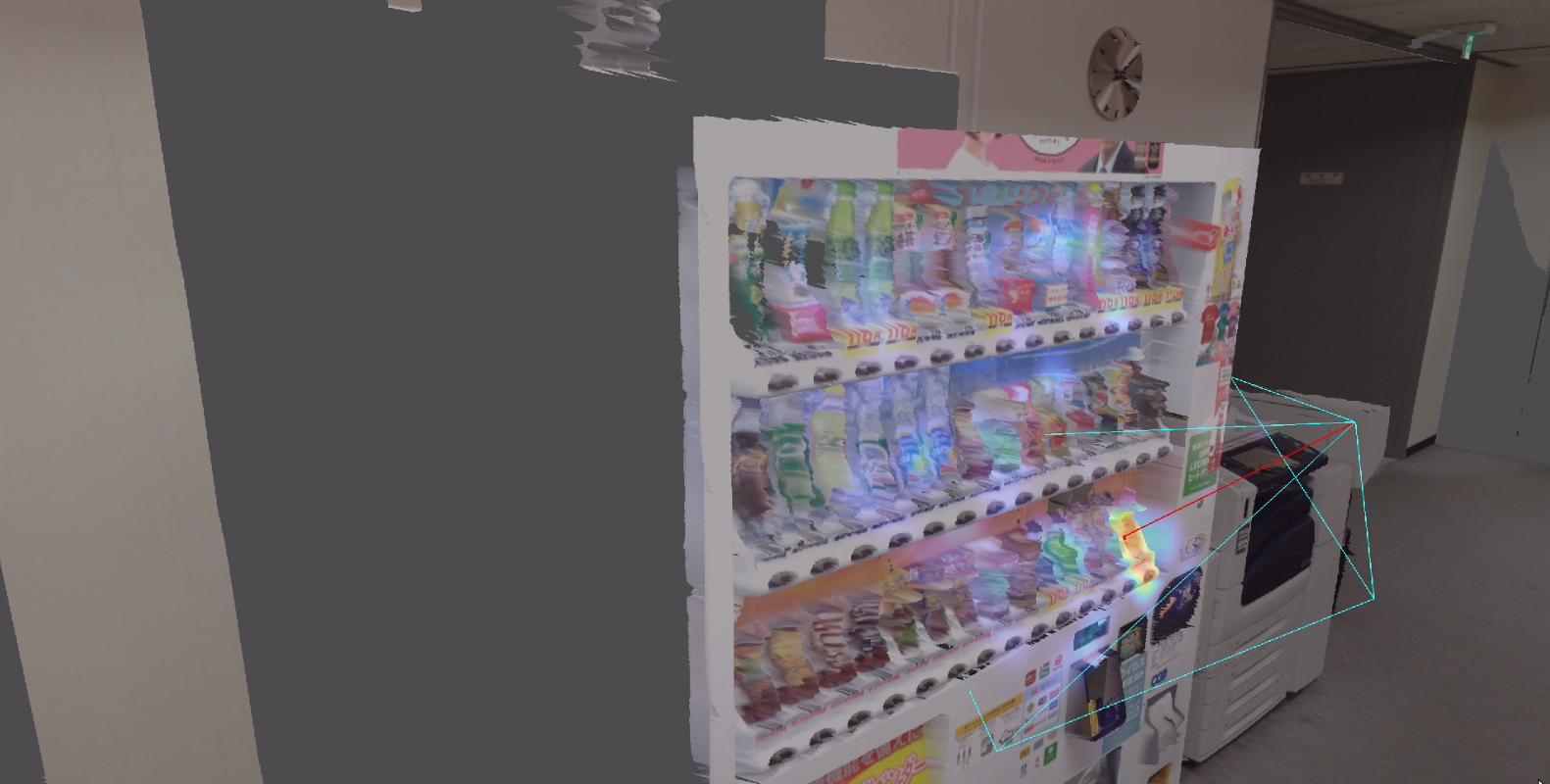} & \includegraphics[width = 0.24\linewidth]{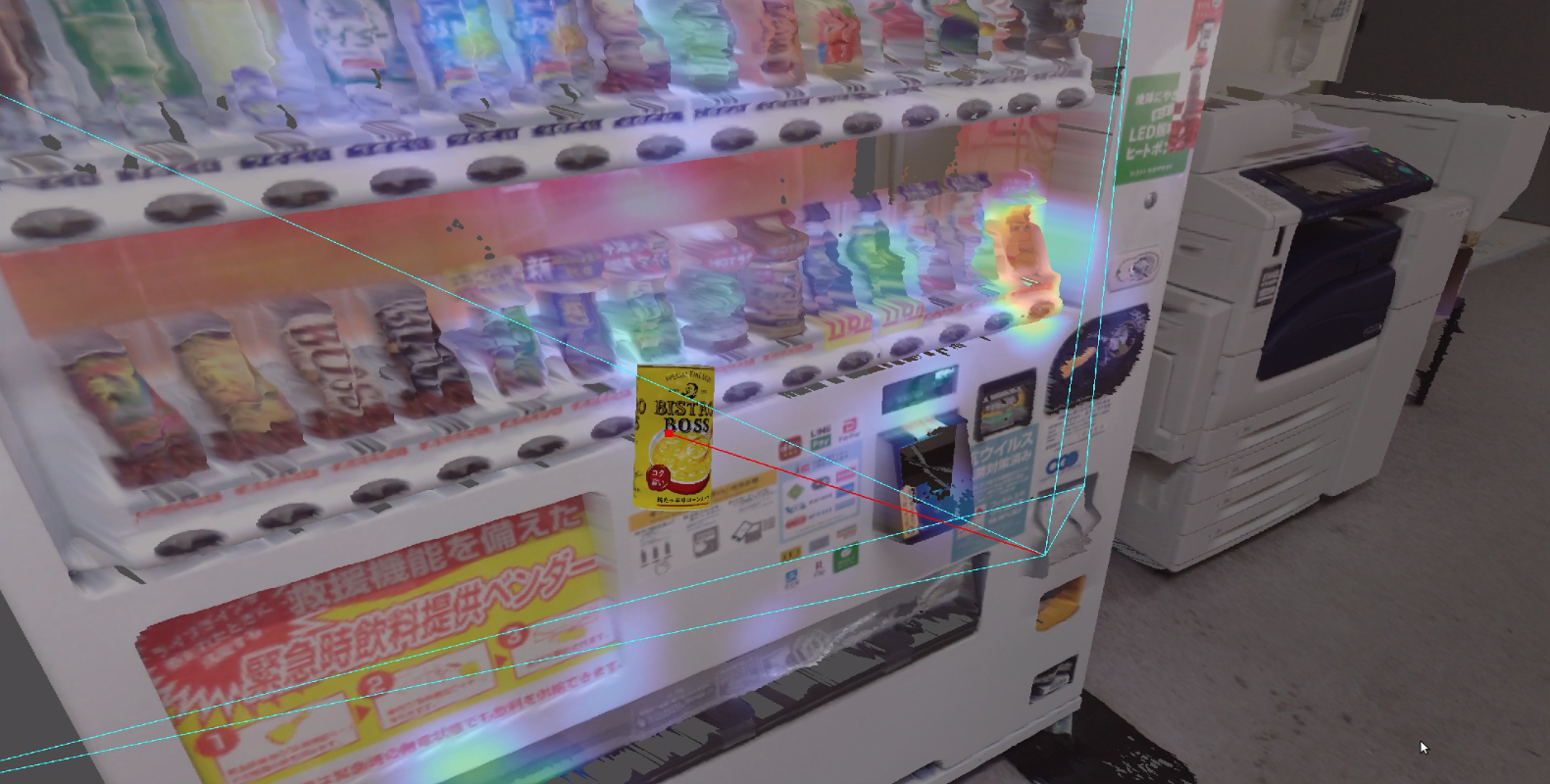} \\
 \end{tabular}}
 \\ (c) Case 3: Pass by a person and buy a drink from a vending machine \vspace{-1mm}

 \end{center}
\vspace{-3mm}
  \caption{First-person views and attention accumulation in different situations. 4D attention successfully localized the subject and simultaneously detected moving objects. Spatio-temporal human attention was accumulated on the target models according to the subject's observations.}
  \label{fig:experiments}
\vspace{-1mm}
 \end{figure}

\begin{figure}[t]
	\scriptsize
	\vspace{2mm}
	\begin{center}
		\includegraphics[width = 0.90\linewidth]{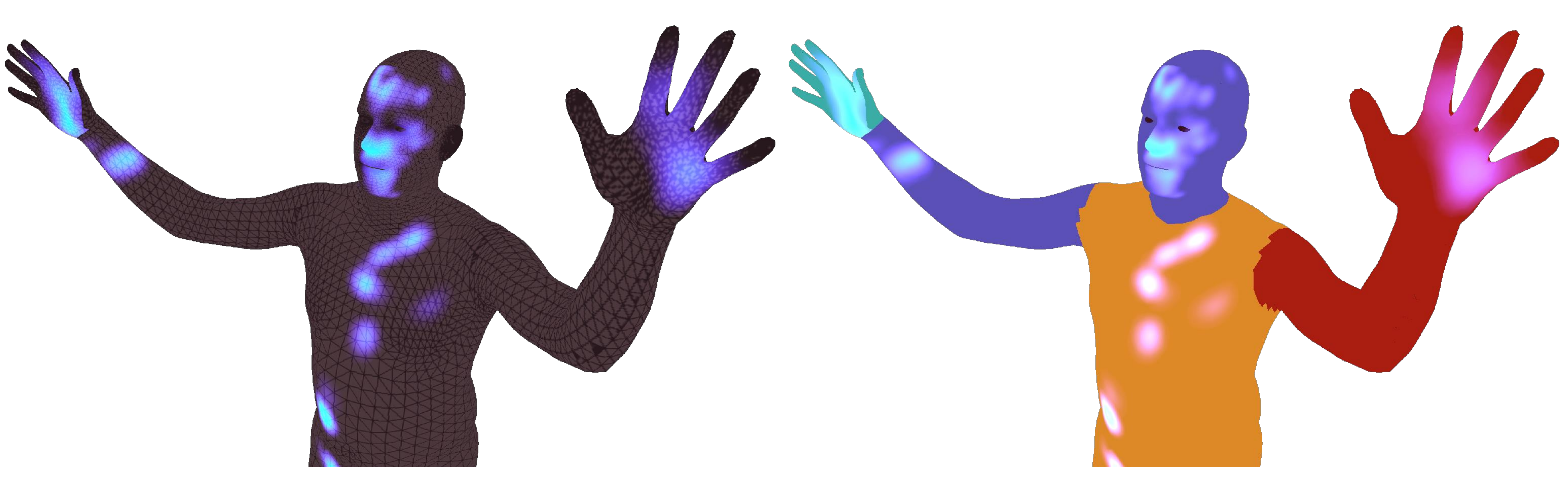} \\
		(a) Human model in Case 1: Appearance (left) and semantics (right) \vspace{2mm} \\
		\includegraphics[width = 0.80\linewidth]{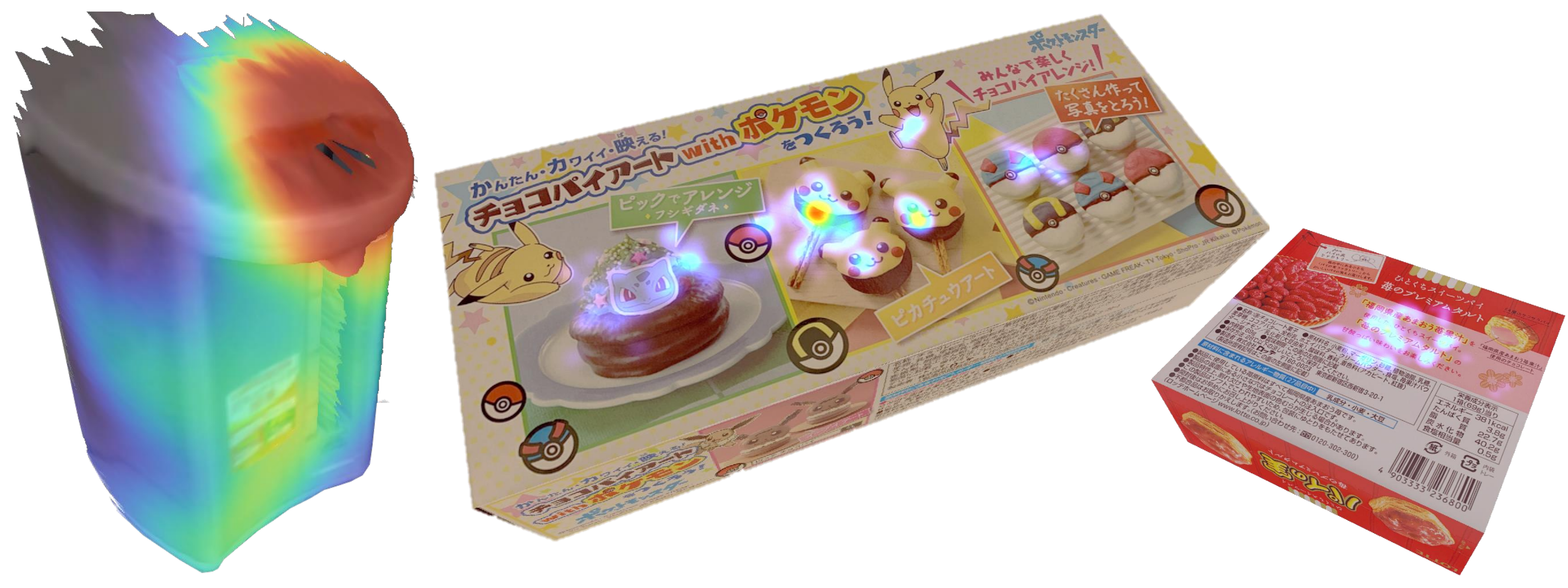} \\
		(b) Electric pot and snacks in Case 2 \vspace{2mm} \\
		\includegraphics[width = 0.80\linewidth]{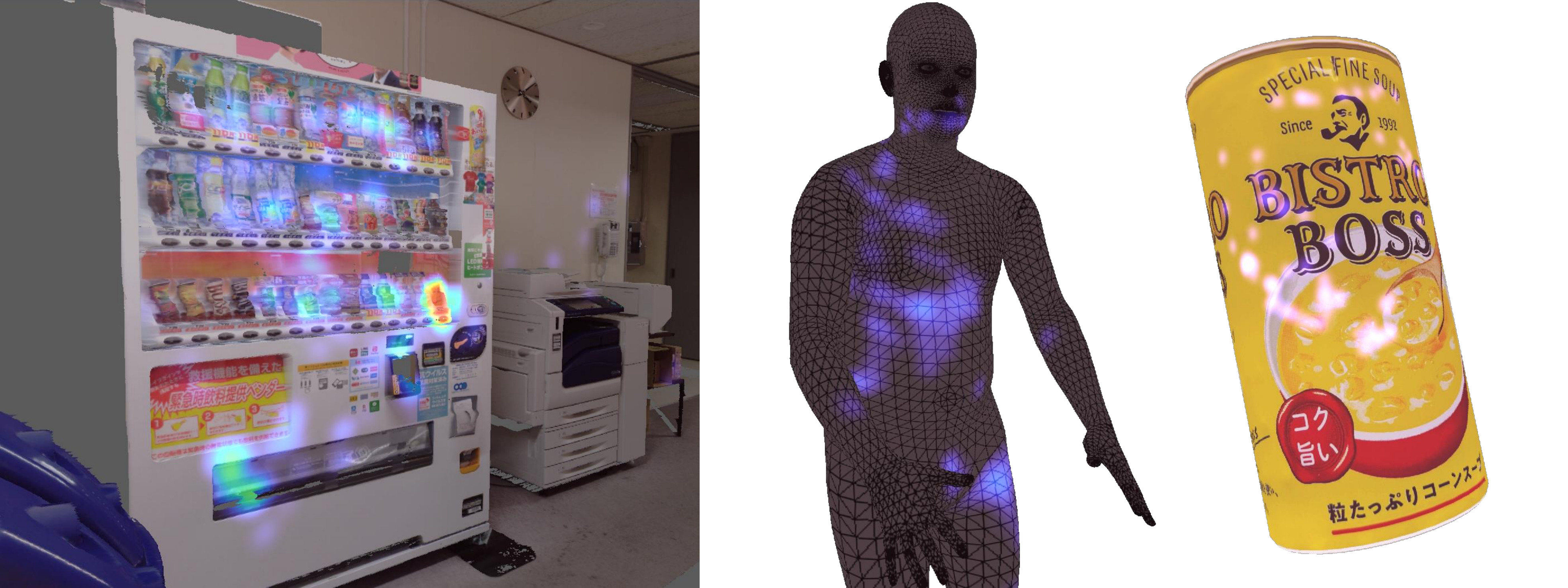} \\
		(c) Vending machine, human, and can models in Case 3 \\
	\end{center}
  \vspace{-3mm}
	\caption{Attractive models in each case with accumulated attention}
	\label{fig:attention}
  \vspace{-3mm}
\end{figure}

\subsection{Applications}
\label{sec:applications}

To further evaluate our method, we performed attention mapping in three realistic situations as shown in Fig. \ref{fig:experiments}.
Figure \ref{fig:attention} picks up ``attractive'' models in each case, in which accumulated human gaze is highlighted.
4D Attention robustly estimated the subject's poses and 3D gaze directions, and simultaneously projected human gaze onto the static and dynamic targets.
This facilitates the analysis of human intention or semantic understanding of the subject's perceptual activities in the real world.

{\bf Case 1:}
As described in Sec.\ref{subsec:gaze_mapping}, attaching different types of textures onto the models makes it possible to access various properties of the models, for example, semantics (see Fig. \ref{fig:attention}(a)).
We easily understand which body parts the subject was focusing on (face and hands, in this case).

{\bf Case 2:}
Instance object reconstruction allows us to observe human attention in highly dynamic situations, for example, object manipulation.
In case 2, after pouring hot water into the mug, the subject picked up freebies and took one.
By accumulating gaze information on the models, we may acquire cues to determine the reason for the subject's choice (Fig. \ref{fig:attention}(b)).

{\bf Case 3:}
We simulated a more realistic situation: The subject walked to a vending machine passing by a person and bought a drink from it.
Our method successfully provided the trajectory, and attention to the static and dynamic objects of the subject (Fig. \ref{fig:attention}(c)), which helps in determining human behavior in the spatio-temporal domain.

\section{DISCUSSION}
\label{sec:discussion}

In this section, we discuss the contributions, limitations, and practicality of the proposed method.
According to Table \ref{tab:comparison}, which comprehensively compares the characteristics of different works, our framework is distinguished from other competitive methods in several aspects, for example, various targets, real-time operation, and easy setup on a simple 3D map.
In particular, the rendering-centered framework provides significant benefits to direct localization and gaze projection via ID texture mapping, which leads to high accuracy of attention mapping as demonstrated in the evaluations.

Map-based methods, however, require a denser 3D map for accurate localization and attention mapping, which can also be a limitation of 4D Attention.
Large 3D map reconstruction and rendering can restrict the application of the method to certain scenes.
Fortunately, 3D reconstruction technologies, such as SLAM with LiDAR\cite{Yokozuka-ICRA2021} or RGB-D cameras\cite{Lee-CVPR2020}, have evolved and are widely available.
Techniques such as view frustum culling\cite{Assarsson-JGT2000} also help in rendering large 3D maps for real-time processing for further applications in indoor and outdoor environments.

Moreover, as demonstrated in Section \ref{sec:applications}, learning-based shape inference, for example, \cite{Rong-arXiv2020}\cite{Manhardt-arXiv2020}, enables attention mapping to unknown dynamic objects by reconstructing target shapes on the fly.
This also allows easier setup to free us from 3D modeling of specific objects, and strengthens our framework toward various usages.

\section{CONCLUSIONS}
\label{sec:conclusion}

We developed a novel gaze-mapping framework to capture human attention in the real world.
The experiments demonstrated that the combination of robust camera localization, unified attention mapping, and instant object reconstruction enables access to 4D human attention.

The proposed system is capable of providing a series of human head poses (trajectory) and simultaneous gaze targets; thus, it would be applicable in action recognition, for example, skill-level evaluation in humanitude tender-care \cite{Nakazawa-JIRS2019}.
It also allows us to incorporate any desired components of instance object reconstruction into the framework, which facilitates attention analysis to specific objects and is helpful for gaze-based target selection in dynamic scenes \cite{Chacn-IROSW2018}.
Additionally, gaze accumulation on 3D models with multiple textures enables semantic analysis of human behavior.

\begin{table*}[t]
\vspace{2mm}
\caption{Comparison of gaze mapping methods}
  \label{tab:comparison}
  \vspace{-4mm}
  \fontsize{6pt}{7pt}\selectfont
  \begin{center}
  {\tabcolsep = 1.0mm \begin{tabular}{l|cccc|cccc} \hline
	    method                          & \multicolumn{2}{c}{target}                  & scalable    & Real-time   & sensors except                 & map                                          & localization                     & attention mapping                  \\ \cdashline{2-3}[0.5pt/1pt]
					    & static map & dynamic objects                &             &             & eye tracker                    &                                              &                                  &                                    \\ \hline \hline                                                            
	    \cite{Fathaliyan-Frontiers2018} &            & \checkmark \textdagger         &             & \checkmark  & Motion capture                 & -                                            & Motion capture                   & Ray casting                        \\
	    \cite{Dini-IROS2017}            &            & \checkmark \textdagger         &             & \checkmark  & Motion capture                 & -                                            & Motion capture                   & Ray casting (Sphere approx.)       \\
	    \cite{Maekawa-ICCVW2019}        & \checkmark &                                & \checkmark  &             & LiDAR \& Motion capture \& IMU & 3D point cloud                               & AMCL \& Motion capture           & Exhaustive examination             \\
	    \cite{Paletta-IRCV2013}         & \checkmark &                                & \checkmark  & \checkmark  & RGB camera                     & Color meshes \& feature points\textdaggerdbl & Indirect visual localization     & OBB-Tree                           \\
	    \cite{Pfeiffer-ETRA2016a}       &            & \checkmark \textdagger         &             & \checkmark  & RGB camera (\& Kinect)         & -                                            & Visual markers                   & Ray casting (Box approx.)          \\
	    \cite{Hagihara-AH2018}          & \checkmark &                                & \checkmark  &             & RGB camera                     & Color meshes \& feature points\textdaggerdbl & Structure-from-Motion            & Ray casting                        \\
	    \cite{Matsumoto-MobileHCI2019}  & \checkmark &                                & \checkmark  &             & RGB camera                     & [Simultaneously built]                       & Mult-View Stereo \& Geometry     & Projection onto Delaunay trinagles \\
	    \cite{Qodseya-ECCVW2016}        & \checkmark & \checkmark                     & \checkmark  &             & Stereo camera                  & [Simultaneously built]                       & RGB-D SLAM                       & 3D cluttered points from the depth \\ \hdashline[0.5pt/1pt]
	    Proposed                        & \checkmark & \checkmark(rigid\&non-rigid)   & \checkmark  & \checkmark  & RGB camera (\& IMU)            & Color meshes                                 & Direct visual localization (C*)  & ID texture mapping                 \\ \hline
    \end{tabular}}
  \end{center}
  \vspace{-1mm}
  \textdagger : Optical or visual marker(s) should be associated with each object for pose tracking. \\
  \textdaggerdbl : Construction of an extra feature-point map that is strictly aligned to the 3D map is required for localization.
  \vspace{-4mm}
\end{table*}



\bibliographystyle{IEEEtran}
\bibliography{myrefs_RA-L2021.bib}

\end{document}